\documentclass[manuscript]{acmart}
\usepackage{pgfplots}
\pgfplotsset{compat=1.18}
\usepackage{tabularray}
\usepackage{tikz}
\usetikzlibrary{
    shapes.geometric,
    arrows.meta,
    positioning 
}
\usepackage{tabularx}
\usepackage{multirow}
\usepackage{graphicx}
\usepackage{enumitem}
\definecolor{coloraug}{RGB}{112, 173, 71}  
\definecolor{coloranno}{RGB}{255, 192, 0} 
\definecolor{colorgan}{RGB}{68, 114, 196}   
\definecolor{colorssl}{RGB}{237, 125, 49}  
\newcommand{\gdot}[1]{\tikz\draw[#1, fill=#1] (0,0) circle (2pt);}
\usepackage{float}
\usepackage{booktabs}
\usepackage{ragged2e} 

\AtBeginDocument{%
  }

\setcopyright{acmlicensed}
\copyrightyear{2025}
\acmYear{2025}
\acmDOI{XXXXXXX.XXXXXXX}
\acmConference[Conference acronym 'XX]{Make sure to enter the correct
  conference title from your rights confirmation email}{June 03--05,
  2018}{Woodstock, NY}
\acmISBN{978-1-4503-XXXX-X/2018/06}

\begin{document}

\title{Beyond Human Annotation: Recent Advances in Data Generation Methods for Document Intelligence}


\author{Dehao Ying}
\affiliation{%
  \institution{Wuhan University}
  \city{Wuhan}
  \country{China}}
\email{yingdehao@whu.edu.cn}

\author{Fengchang Yu}
\affiliation{%
  \institution{Wuhan University}
  \city{Wuhan}
  \country{China}}
\email{yufc2002@whu.edu.cn}

\author{Haihua Chen}
\affiliation{%
  \institution{University of North Texas}
  \city{Denton}
  \state{Texas}
  \country{United States}}
\email{haihua.chen@unt.edu}

\author{Changjiang Jiang}
\affiliation{%
  \institution{Wuhan University}
  \city{Wuhan}
  \country{China}}
\email{jiangcj@whu.edu.cn}

\author{Yurong Li}
\affiliation{%
  \institution{Wuhan University}
  \city{Wuhan}
  \country{China}}
\email{annalee218@whu.edu.cn}

\author{Wei Lu}
\affiliation{%
  \institution{Wuhan University}
  \city{Wuhan}
  \country{China}}
\email{weilu@whu.edu.cn}

\renewcommand{\shortauthors}{YING et al.}

\begin{abstract}
The advancement of Document Intelligence (DI) demands large-scale, high-quality training data, yet manual annotation remains a critical bottleneck. While data generation methods are evolving rapidly, existing surveys are constrained by fragmented focuses on single modalities or specific tasks, lacking a unified perspective aligned with real-world workflows. To fill this gap, this survey establishes the first comprehensive technical map for data generation in DI. Data generation is redefined as supervisory signal production, and a novel taxonomy is introduced based on the "availability of data and labels." This framework organizes methodologies into four resource-centric paradigms: \textbf{Data Augmentation}, \textbf{Data Generation from Scratch}, \textbf{Automated Data Annotation}, and \textbf{Self-Supervised Signal Construction}. Furthermore, a multi-level evaluation framework is established to integrate \textbf{intrinsic quality} and \textbf{extrinsic utility}, compiling performance gains across diverse DI benchmarks. Guided by this unified structure, the methodological landscape is dissected to reveal critical challenges such as fidelity gaps and frontiers including co-evolutionary ecosystems. Ultimately, by systematizing this fragmented field, data generation is positioned as the central engine for next-generation DI.
\end{abstract}

\ccsdesc[500]{General and reference~Surveys and overviews}
\ccsdesc[500]{Computing methodologies~Artificial intelligence}

\keywords{Document Intelligence, Data Generation, Data Quality Evaluation}


\maketitle

\section{INTRODUCTION}
Document Intelligence (DI), an interdisciplinary field integrating computer vision, natural language processing, and pattern recognition, aims to enable machines to ``read'' and ``understand'' documents\cite{cui2021document, ke2025large}. It spans a wide range of tasks from low-level perception, such as Layout Analysis, to high-level cognition, like Document Visual Question Answering (DocVQA). Progress in this field is currently driven by two main technological waves. The first is the rise of specialized document foundation models based on the Transformer architecture, including LayoutLM\cite{xu2020layoutlm}, DiT\cite{li2022dit}, Donut\cite{kim2022ocr}, and UDOP\cite{tang2023unifying}. The second is the growing adoption of general-purpose Large Language Models (LLMs) for DI tasks. The success of both approaches\textemdash whether through large-scale pre-training of specialized models or efficient instruction tuning of general-purpose LLMs—relies heavily on vast quantities of high-quality training data.

\begin{figure*}[t]
  \centering
  \includegraphics[width=\textwidth]{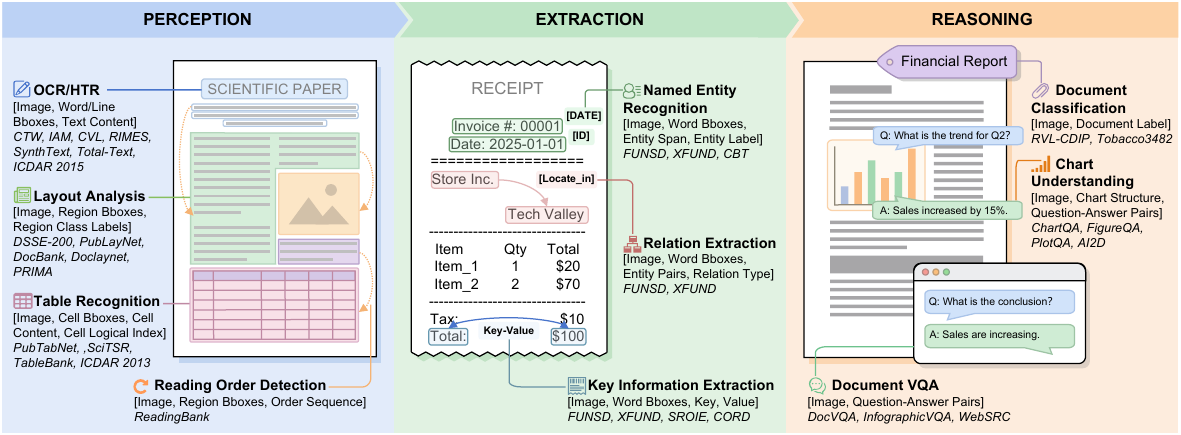}
  \caption{Overview of Representative Document Intelligence Tasks, Their Required Data Structures, and Benchmarks.}
  \label{fig:Figure1}
\end{figure*}

However, the data requirements for DI are far more stringent than those for general CV or NLP tasks, distinguished by three key characteristics: (1) Inherent Multimodality: Data is inherently multimodal, where meaning emerges from the interplay of visual, textual, and layout cues. (2) Fine-Grained, Task-Specific Annotations: Tasks demand intricate labels, such as assigning semantic categories, text content, and word-level bounding boxes for Key Information Extraction (KIE) and creating coherent document-question-answer triplets for DocVQA. (3) High Demand for Diversity: To ensure robust generalization, training data must cover a vast spectrum of real-world documents, varying in type, layout, language, and quality. To better illustrate these characteristics and provide necessary context for the data generation techniques discussed in subsequent sections, we organize the core DI tasks into three levels based on the cognitive depth: Document Parsing, Information Extraction, and Document Understanding and Reasoning. Figure~\ref{fig:Figure1} details these representative tasks, showing their specific training data structures and their common datasets.

\begin{figure}[t]
    \centering
    \definecolor{nodecolor}{RGB}{31, 119, 180}  
    \begin{tikzpicture}
        \begin{axis}[
            axis lines=left,
            xlabel={Year},
            xlabel style={
                at={(rel axis cs:1,0)}, 
                anchor=north west,       
                xshift=2pt,              
                yshift=-2pt              
            },
            ylabel={Number of Publications},
            ymin=0, ymax=350,
            xmin=2015.5, xmax=2025.5,
            xtick={2016, 2017, 2018, 2019, 2020, 2021, 2022, 2023, 2024, 2025},
            xticklabel style={
                yshift=-5pt,
            },
            ytick={0, 100, 200, 300},
            ymajorgrids=true,
            grid style={
                color=gray,
                opacity=0.3,
                dashed
            },
            width=0.95\textwidth,
            height=5cm,
            enlarge x limits=0.05,
            label style={font=\footnotesize},
            tick label style={font=\footnotesize},
        ]
            \addplot[
                color=nodecolor,
                mark=*,
                mark options={fill=nodecolor, scale=0.9},
                thick,
                nodes near coords,
                nodes near coords align={above},
                node near coords style={font=\footnotesize, color=black},
            ] 
            coordinates {
                (2016, 36)
                (2017, 56)
                (2018, 60)
                (2019, 90)
                (2020, 139)
                (2021, 164)
                (2022, 192)
                (2023, 208)
                (2024, 286)
                (2025, 285)
            };
        \end{axis}
    \end{tikzpicture}
    \caption{The trend of publications on data generation for document intelligence over the past decade. Data is retrieved from Scopus using the query: \texttt{TITLE-ABS-KEY((``document'') AND (``data augmentation'' OR ``data synthesis'' OR ``data generation'' OR ``synthetic data'' OR ``artificial data''))}, filtered for publication years 2016--2025.}
    \label{fig:publication_trend}
\end{figure}
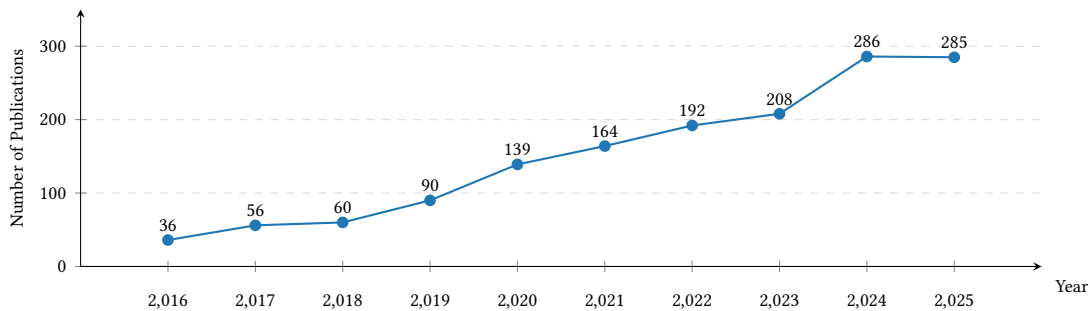

Given these stringent data requirements, public datasets are often insufficient due to their limited scale, domain specificity, or annotations ill-suited for certain tasks. Manual annotation has also become a significant bottleneck, limited by prohibitive costs and time due to its labor-intensive nature, a heavy reliance on domain experts for specialized documents, and an inability to cover long-tail or privacy-sensitive scenarios. These challenges underscore the urgent need for methods that go \textbf{Beyond Human Annotation} to acquire large-scale, high-quality training data in an efficient, cost-effective, and secure manner. In response, a research field centered on data generation has emerged and is rapidly growing. This rapid growth is corroborated by publication trends. A search on Scopus shows that the volume of related papers from 2021–2025 has almost tripled compared to the 2016–2020 period, as shown in Figure~\ref{fig:publication_trend}.

However, this expansion has led to varying interpretations of what constitutes ``data generation.'' To provide a clear boundary for this review, we explicitly define the \textbf{scope} along two dimensions:
First, regarding the \textit{methodology}, we use ``data generation'' as an umbrella term to encompass all non-manual methods for programmatically creating, augmenting, or automatically producing supervisory signals. Consequently, strategies relying solely on human effort\textemdash such as crowdsourcing, manual annotation interface optimization, or active learning (sample selection without generation)\textemdash fall outside our scope.
Second, regarding the \textit{domain}, we focus exclusively on Visually-Rich Documents (VRDs) where semantic understanding relies on the interplay of visual, text and layout features. Therefore, methods targeting pure plain text or natural scene text (e.g., street signs) are excluded unless explicitly adapted for document structures.

While numerous surveys on data generation exist in the general CV and NLP fields, they face critical limitations when applied to DI, as summarized below:

\begin{itemize}
    \item \textbf{Domain Specificity:} No comprehensive survey specifically addresses data generation methods for the holistic field of DI. Existing surveys are often fragmented, focusing either on a single modality (e.g., text data augmentation ~\cite{chai2026text}) or a specific downstream task (e.g., text classification~\cite{bayer2022survey}). The former can not capture the intrinsic multimodal vision-text-layout nature of documents, while the latter creates information silos that hinder the transfer of effective generation techniques across related tasks.
    
    \item \textbf{Taxonomy Rationale:} Existing comprehensive surveys are predominantly organized by technique~\cite{shorten2019survey}. While theoretically sound, a technique-based classification is practically misaligned with the decision-making process. It offers little guidance for addressing fundamental resource-constrained questions like, ``Given a collection of unlabeled scans, how can I leverage them?'' or ``How can I support a new type of invoice from scratch?'', as it scatters solutions for the same resource scenario across disparate technical chapters.
    
    \item \textbf{Scope Completeness:} Existing surveys have a narrow definition of ``data generation,'' almost universally confining it to the creation of labeled training samples for downstream supervised models, while completely ignoring the critical paradigms of ``automated annotation'' and ``self-supervised signal construction.''
    
    \item \textbf{Evaluation Depth:} Existing surveys generally lack a systematic, multi-level evaluation framework for the quality of generated data, often simply equating downstream task performance with intrinsic data quality.
\end{itemize}

To address these issues, this survey introduces a novel, resource-centric and task-agnostic taxonomy for data generation methods. The taxonomy centers on the ``availability of data and labels,'' providing a unified decision map for technology selection across diverse data resource scenarios, and divides all methods into four categories: (1) \textbf{Data Augmentation}, for expanding a dataset when both samples and labels are available; (2) \textbf{Data Generation from Scratch}, for creating new, labeled data when neither samples nor labels are available; (3) \textbf{Automated Data Annotation}, for accelerating the production of high-quality labels when only unlabeled samples exist; and (4) \textbf{Self-Supervised Signal Construction}, for automatically creating training signals for pre-training a foundation model when abundant unlabeled samples are available. We deliberately organize the main text by these resource-based paradigms rather than by task to break down information silos and promote technological transferability. This structure enables researchers to identify cross-domain opportunities—such as adapting a technique originally used in OCR to improve KIE—based on their specific resource constraints. Conversely, since benchmarks are inherently task-specific, we organize the extrinsic quantitative evaluation by downstream task. This orthogonal design ensures that the methodology sections elucidate the generation mechanisms across different resource settings, while the evaluation tables rigorously verify their utility on standardized benchmarks. Finally, we integrate a systematic framework for assessing intrinsic quality, providing a ready-to-use evaluation toolbox for the field.

The main contributions of this survey are:
\begin{itemize}
    \item The first comprehensive review of data generation for document intelligence.
    \item An expanded definition of ``data generation'' that moves beyond sample creation to include automated annotation and self-supervised signal construction.
    \item A novel, resource-centric and task-agnostic taxonomy based on the ``availability of data and labels'' that facilitates cross-task methodological transfer and provides practical guidance for technology selection.
    \item A systematic evaluation framework assessing both the intrinsic quality of synthetic data and its extrinsic utility on downstream task benchmarks.
\end{itemize}

The remainder of this survey is organized as follows. Section~\ref{sec:taxonomy} details our novel taxonomy. Subsequently, Sections~\ref{sec:data_augmentation}–\ref{sec:self_supervised} provide systematic reviews of the techniques corresponding to the four paradigms. Section~\ref{sec:evaluation} establishes the comprehensive evaluation framework. Finally, we discuss insights and trends in Section~\ref{sec:discussion} and conclude in Section~\ref{sec:conclusion}.

\section{TAXONOMY}
\label{sec:taxonomy}
\subsection{Problem Formulation and Dimensions}
Data generation in DI is fundamentally a process of acquiring supervisory signals under specific resource constraints and learning objectives. Formally, let $\mathcal{X}$ denote the document space and $\mathcal{Y}$ denote the target label space. The goal is to obtain a training set $\mathcal{D}$ derived from an initial resource set $\mathcal{R}$. 

\begin{figure}[t]
    \centering
    \definecolor{startColor}{RGB}{149, 149, 149}
    \definecolor{decisionColor}{RGB}{200, 200, 255}
    \definecolor{branch1Color}{RGB}{193, 225, 193}  
    \definecolor{branch2Color}{RGB}{31, 119, 180}   
    \definecolor{branch3Color}{RGB}{255, 247, 188}  
    \definecolor{branch4Color}{RGB}{254, 204, 153}  

    \begin{tikzpicture}[
        node distance=0.8cm and 1.2cm 
    ]
        \tikzset{
            base/.style = {
                draw, 
                thick, 
                align=center, 
                font=\sffamily\footnotesize,
                minimum height=1.2cm
            },
            startNode/.style = {
                base, 
                rectangle, 
                rounded corners=1pt, 
                fill=startColor!25, 
                draw=startColor!80!black, 
                minimum width=2.5cm,
                minimum height=1cm
            },
            decisionNode/.style = {
                base, 
                diamond, 
                aspect=1.8, 
                fill=decisionColor!40, 
                draw=decisionColor!70!black, 
                inner sep=0pt, 
                minimum width=2.5cm
            },
            branchNode/.style = {
                base, 
                rectangle, 
                rounded corners=3pt, 
                minimum width=2.5cm,
                minimum height=1cm
            },
            branch1/.style={branchNode, fill=branch1Color!50, draw=branch1Color!70!black},
            branch2/.style={branchNode, fill=branch2Color!20, draw=branch2Color!90}, 
            branch3/.style={branchNode, fill=branch3Color!60, draw=branch3Color!80!black}, 
            branch4/.style={branchNode, fill=branch4Color!50, draw=branch4Color!70!black},
            arrowLabel/.style = {font=\sffamily\footnotesize, sloped}
        }

        \node (start) [startNode] {Start a Document \\ Intelligence Project};
        
        \node (dec1)  [decisionNode, right=0.5cm of start] {Existing \\ Samples?};
        
        \node (b2)    [branch2, below right=0.4cm and 0.8cm of dec1] {Branch 2: \\ Data Generation \\ from Scratch};

        \node (dec2)  [decisionNode, above right=0.4cm and 1.4cm of dec1] {Existing \\ Annotations?};
        
        \node (b1)    [branch1, above right=0.4cm and 0.8cm of dec2] {Branch 1: \\ Data Augmentation};
        
        \node (dec3)  [decisionNode, below right=0.4cm and 1.4cm of dec2] {Supervised \\ Learning?};
        
        \node (b3)    [branch3, above right=0.3cm and 0.8cm of dec3] {Branch 3: \\ Automated Data \\ Annotation};
        
        \node (b4)    [branch4, below right=0.4cm and 0.8cm of dec3] {Branch 4: \\ Self-Supervised \\ Signal Construction};

        \draw[->, >=Stealth] (start) -- (dec1);
        
        \draw[->, >=Stealth] (dec1.south) to[out=-45, in=180] node[arrowLabel, above] {No} (b2.west);
        
        \draw[->, >=Stealth] (dec1.north) to[out=30, in=180]  node[arrowLabel, above] {Yes} (dec2.west);
        \draw[->, >=Stealth] (dec2.north) to[out=45, in=180]  node[arrowLabel, above] {Yes} (b1.west);
        \draw[->, >=Stealth] (dec2.south) to[out=-30, in=180] node[arrowLabel, above] {No} (dec3.west);
        
        \draw[->, >=Stealth] (dec3.north) to[out=30, in=180]  node[arrowLabel, above] {Yes} (b3.west);
        \draw[->, >=Stealth] (dec3.south) to[out=-45, in=180] node[arrowLabel, above] {No} (b4.west);
        
    \end{tikzpicture}
    \caption{The decision flowchart for the taxonomy, guiding paradigm selection based on resource constraints and learning objectives.}
    \label{fig:taxonomy_flowchart}
\end{figure}
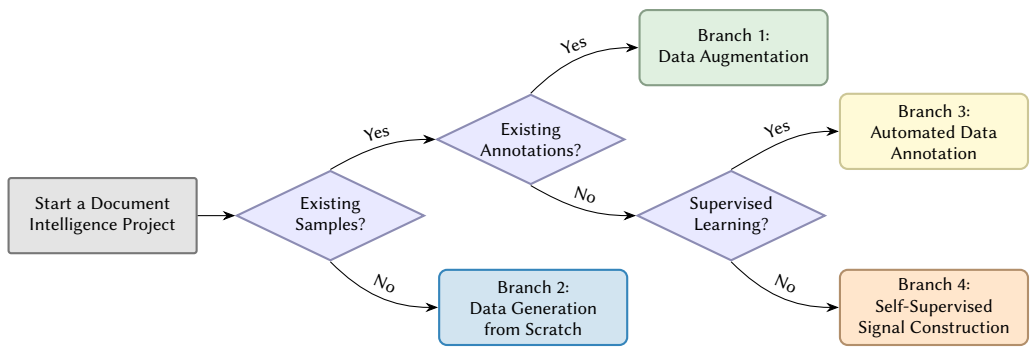

Unlike conventional frameworks organized simply by technique, our taxonomy is \textbf{resource-centric and objective-driven}. We classify data generation paradigms by traversing a decision path defined by three logical dimensions, as illustrated in the decision flowchart in Figure~\ref{fig:taxonomy_flowchart}:
\begin{enumerate}
    \item \textbf{Data Availability:} Is raw document data ($x \in \mathcal{X}$) available?
    \item \textbf{Label Availability:} Is ground-truth annotation ($y \in \mathcal{Y}$) available?
    \item \textbf{Target Learning Paradigm:} Is the goal to construct explicit input-output pairs for \textit{Supervised Learning}, or to construct pretext tasks for \textit{Self-Supervised Learning}?
\end{enumerate}

\subsection{Four Paradigms of Data Generation}
Based on the intersection of these dimensions, we categorize methodologies into four paradigms. The first three paradigms share a common objective: to construct or expand a labeled dataset $\mathcal{D} = \{(x, y)\}$ to enable or enhance downstream \textbf{Supervised Learning}. They differ primarily in their starting resource constraints ($\mathcal{R}$).

\textbf{1. Data Augmentation (Data $\checkmark$, Label $\checkmark$, Goal: Supervised).} 
This paradigm applies when a labeled dataset $\mathcal{R} = \{(x, y)\}$ is already available but may be limited in size or diversity. The core mechanism is to apply a label-preserving transformation function $\mathcal{T}$ to generate new samples $(x', y) = \mathcal{T}(x, y)$. This process enhances model robustness and generalization without altering the core semantics or label correctness.

\textbf{2. Data Generation from Scratch (Data $\times$, Label $\times$, Goal: Supervised).} 
Addressing the cold-start problem where neither representative samples nor labels are available ($\mathcal{R} = \emptyset$), this paradigm synthesizes completely new pairs $(x, y)$ from abstract templates or noise. It involves algorithms that simultaneously generate the visual appearance of a document and its precise structured labels, fundamentally solving data privacy or scarcity issues.

\textbf{3. Automated Data Annotation (Data $\checkmark$, Label $\times$, Goal: Supervised).} 
This paradigm addresses the scenario where abundant unlabelled samples $\mathcal{R} = \{x\}$ exist. It aims to approximate the mapping $y = \mathcal{F}(x)$ using programmatic rules, specialized teacher models, or LLMs. This transforms the laborious annotation task from ``creation'' to ``verification,'' efficiently producing high-quality labels for downstream supervision.

Finally, faced with the same resource condition as Automated Data Annotation (abundant unlabelled data), the fourth paradigm diverges in its learning objective.

\textbf{4. Self-Supervised Signal Construction (Data $\checkmark$, Label $\times$, Goal: Self-Supervised).} 
Instead of generating explicit labels for specific downstream tasks, this paradigm mines intrinsic supervision. It constructs pretext tasks where a pseudo-label $\tilde{y}$ is derived from the data itself $(x, \tilde{y}) = \mathcal{P}(x)$. The goal is \textbf{Representation Learning}—allowing foundation models to learn transferable features by ``teaching themselves.''

Together, these four categories form a complete map of data generation in DI, covering all scenarios from enhancing existing data to creating new data, and from serving downstream tasks to empowering upstream foundation models. The subsequent sections will provide a detailed review of the key techniques, representative works, and trade-offs within each category. Furthermore, we systematically investigate the evaluation methodologies for these techniques and compare their quantitative performance across widely adopted DI benchmarks.

\section{DATA AUGMENTATION}
\label{sec:data_augmentation}
\begin{figure*}[t]
  \centering
  \includegraphics[width=\textwidth]{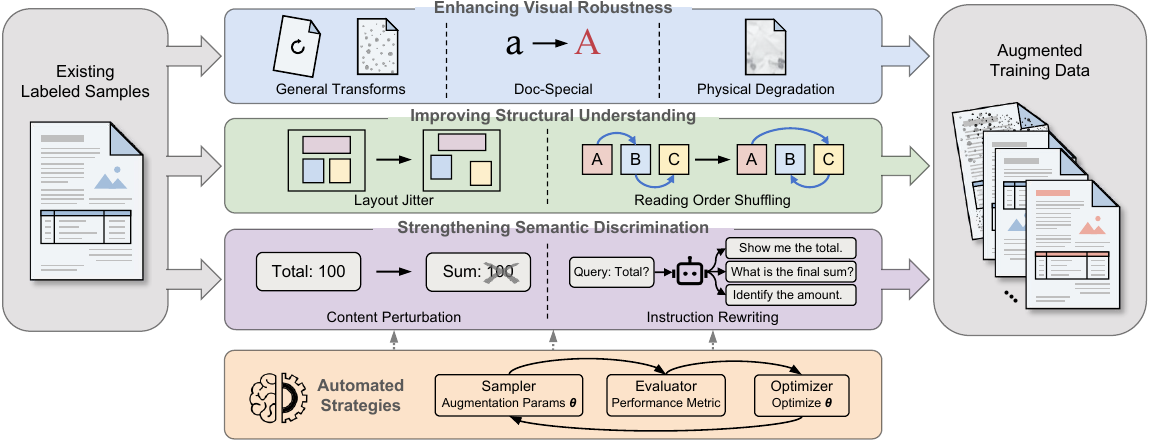}
  \caption{A unified framework for Data Augmentation. The methods are organized hierarchically into four layers based on the augmentation objective: (1) Enhancing Visual Robustness via geometric and physical transformations; (2) Improving Structural Understanding by perturbing layout and reading order; (3) Strengthening Semantic Discrimination through content modification and instruction rewriting; and (4) Automated Strategies that dynamically optimizes augmentation policies.}
  \label{fig:Figure4}
\end{figure*}

The evolution of data augmentation reflects a deepening understanding of the multimodal nature of documents. Accordingly, this section organizes these techniques based on their core objectives, as illustrated in Figure~\ref{fig:Figure4}. Section~\ref{sec:visual_robustness} explores \textbf{Enhancing Visual Robustness}, which treats documents as visual objects and applies visual and physical transformations. Section~\ref{sec:structural_understanding} delves into \textbf{Improving Structural Understanding}, which views documents as structured objects and perturbs their layout, order, or annotations. Section~\ref{sec:semantic_discrimination} examines \textbf{Strengthening Semantic Discrimination}, which considers documents as semantic entities and modifies their text content. Finally, Section~\ref{sec:automated_augmentation} introduces \textbf{Automated Augmentation Strategies}, which aim to automate the augmentation process.

\subsection{Enhancing Visual Robustness}
\label{sec:visual_robustness}
Recent benchmarks have revealed that even state-of-the-art DI models are remarkably fragile to out-of-distribution visual shifts~\cite{he2023good, chen2024rodla}. By introducing perturbations such as background replacement or simulating real-world processing distortions, these works demonstrate a stark gap between model performance on clean versus noisy data. This fragility underscores the critical need for visual augmentation to build truly practical systems.

\subsubsection{General Geometric and Optical Transformations}
Geometric (e.g., rotation, scaling) and optical (e.g., blur, noise) transformations serve as the foundational, low-cost baseline for injecting diversity. Their effectiveness is well-established across tasks ranging from OCR~\cite{li2023trocr}, table detection~\cite{cherepanov2020automated}, KIE~\cite{amari2024efficient}, and even in the pre-training of large foundation models~\cite{powalski2021going}.
Beyond their direct application, these transformations are also crucial in Data Generation from Scratch (see Section~\ref{sec:generation_from_scratch}). They are used as post-processing steps to enhance the realism of rendered outputs, such as handwritten text or ID cards~\cite{pippi2023handwritten, monsur2023synthnid}. More creatively, they are used as pre-processing steps, augmenting foundational elements like tables and figures before composing the final document to enrich the diversity of the raw materials~\cite{zhao2024doclayout}.

\subsubsection{Document-Specific Visual Transformations}
Researchers have also designed augmentation strategies tailored to the unique typographic and structural properties of documents. At the micro-level, efforts focus on diversifying \textbf{text appearance}. This involves randomizing fonts, styles, and colors during the rendering process~\cite{kim2023web, aggarwal2023dublin}, or directly parsing and modifying font attributes (e.g., bold, italics) within existing PDF files~\cite{kumar2024gridstemlaynet}. Some approaches also simulate capture artifacts, such as mimicking camera shake by randomly resizing image patches~\cite{shi2023multi}. Moving to the element-level, specific transformations target \textbf{structural components}. For instance, CascadeTabNet~\cite{prasad2020cascadetabnet} introduces table-specific augmentations, including morphological dilation to thicken cell borders and smearing effects to simulate ink bleed.

\subsubsection{Systematic Physical Degradation Simulation}
To realistically simulate the physical artifacts documents accumulate (e.g., from printing, scanning, or handling), some works systematically combine multiple transformations into complete pipelines. Several approaches convert digitally-born papers into pseudo-scanned documents by applying a suite of transformations, such as geometric distortions, noise, and morphological operations~\cite{kahu2021scanbank, blecher2024nougat, sun2024locr, kim2021donut}.

Notably, Augraphy~\cite{groleau2023augraphy} provides a comprehensive Python library that simulates a complex, physics-based document degradation pipeline. Its capabilities include modeling ink effects (e.g., \texttt{InkBleed}), scanner artifacts (e.g., \texttt{DirtyScreen}), and physical damage (e.g., \texttt{Wrinkles}). Recent work~\cite{dai2025enhancing} has demonstrated the library's effectiveness in transforming digital documents into handwritten-like scan images, successfully bridging the domain gap.

\subsection{Improving Structural Understanding}
\label{sec:structural_understanding}
Structural perturbations move beyond pixel-level operations to target the intrinsic ``grammar'' of documents—their layout and reading order. By introducing noise to these structural priors, such strategies force models to learn robust representations of organizational logic, thereby reducing sensitivity to complex layouts or OCR sequencing errors.

These structural perturbations can be applied at different levels. To \textbf{simulate reading order errors}, several works perturb the sequence of document elements, for instance, by globally shuffling all text blocks for severe misalignments or swapping adjacent blocks for minor ones~\cite{wang2022simple, he2023good}. To \textbf{improve tolerance to fine-grained typesetting variations}, another approach scales the horizontal and vertical distances between tokens by a random factor~\cite{powalski2021going}. This principle can also be applied directly to the annotations themselves. For example, LOCR~\cite{sun2024locr} adds Gaussian noise to bounding box coordinates, simulating the imprecise localization or loose boxes common in manual annotation.

\subsection{Strengthening Semantic Discrimination}
\label{sec:semantic_discrimination}
The highest level operates on the semantic content of documents. By modifying text or task instructions, these methods aim to enhance robustness to linguistic variations, mitigate semantic biases, or teach complex instruction-following.

\subsubsection{Classic Text Content Perturbation}
Classic augmentation techniques from NLP are widely adopted to introduce controlled semantic noise. These include methods such as synonym replacement ~\cite{sun2024synonym} and random token deletion to simulate OCR errors~\cite{sun2024locr}. To expand small datasets, \cite{zhang2021entity} also created pseudo-documents by randomly dropping words from entity phrases.

\subsubsection{Creative Semantic Content Generation}
More advanced approaches perform complex, creative modifications on semantic content to address deeper issues. A key application is mitigating model biases. Observing that models often over-rely on language priors rather than visual evidence, MixTex~\cite{luo2024mixtex} created a hybrid dataset with syntactically plausible but semantically incorrect pseudo-elements (e.g., fake formulas) to force a closer reliance on visual details. With the rise of LLMs, another powerful paradigm is to generatively augment task instructions or answers. This includes expanding concise answers into full sentences~\cite{feng2024docpedia}, generating diverse text for entities~\cite{Wojcik2025NewPI}, or creating multiple phrasings of a question for the same image-answer pair~\cite{tanaka2024instructdoc}.

\subsection{Automated Augmentation}
\label{sec:automated_augmentation}
While effective, the aforementioned methods rely on manually defined, fixed transformation rules that may not transfer well across datasets or tasks. To overcome this, automated augmentation paradigms empower algorithms to learn optimal strategies. These approaches range from data-driven heuristics to learning-based policy search.

A straightforward approach is to use the dataset's own statistics to guide the process. For example, TabRecSet~\cite{yang2023large} uses a data-driven probabilistic model for its table structure augmentation, sampling a target dimension based on the global distribution of row and column counts to ensure the augmented tables remain representative. 

A more advanced paradigm formalizes policy selection as a search problem solved via optimization. 
FgAA~\cite{chen2025fine} employs Bayesian optimization to find optimal stroke-level parameters for HTR. It forms a closed loop where the model's validation performance on augmented data guides the optimizer to discover a tailor-made augmentation recipe.

\subsection{Observations}

\begin{table}[t]
\caption{Summary of Data Augmentation Paradigms.}
\label{tab:aug_summary}
\footnotesize
\renewcommand{\arraystretch}{1.3} 
\begin{tabularx}{\textwidth}{
  >{\raggedright\arraybackslash}p{4.2cm} 
  >{\raggedright\arraybackslash}p{3cm}
  >{\raggedright\arraybackslash}p{3cm}         
  >{\raggedright\arraybackslash}X         
}
\hline
\textbf{Category (Techniques)} & \textbf{Advantages} & \textbf{Limitations} & \textbf{Typical Use Cases} \\
\hline
\multicolumn{4}{l}{\textbf{1. Enhancing Visual Robustness}} \\
\hline
\textbf{General Visual Transformations}: \newline (Rotation, scaling, noise, blur)
& Simple, low-cost, universally applicable
& Limited realism for document-specific artifacts
& Baseline augmentation for all tasks \\

\textbf{Doc-Specific Transformations}: \newline (Font randomization, document elements transformation)
& Highly targeted, simulates unique visual properties of documents
& Can be complex to design
& Tasks involving font/style variation or small element perception \\

\textbf{Systematic Degradation}: \newline (Augraphy, scanning simulation)
& High realism, effectively bridges the domain gap
& Computationally intensive, requires specialized tools
& Applications deployed on low-quality scans \\
\hline
\multicolumn{4}{l}{\textbf{2. Improving Structural Understanding}} \\
\hline
\textbf{Structural Perturbations}: \newline (Element shuffling, coordinate noise)
& Enhances robustness to layout shifts and OCR errors
& High risk of corrupting the logical structure
& Layout-sensitive tasks (e.g., KIE, reading order) \\
\hline
\multicolumn{4}{l}{\textbf{3. Strengthening Semantic Discrimination}} \\
\hline
\textbf{Classic Content Perturbations}: \newline (Synonym replacement, token deletion)
& Increases linguistic diversity
& High risk of altering key semantics or labels
& Scenarios requiring improvement in semantic generalization \\

\textbf{Creative Semantic Generation}: \newline (Adversarial samples creation, LLM augmentation)
& Generate high-quality or even adversarial signals
& Design is complex, often relies on powerful LLMs
& Training instruction following models, mitigating model biases \\
\hline
\multicolumn{4}{l}{\textbf{4. Automated Augmentation}} \\
\hline
\textbf{Automated Strategies}: \newline (Data-driven guidance, policy search)
& Adaptive, automatically find optimal policies
& Computationally expensive, lacks interpretability
& Scenarios requiring maximum performance \\
\hline
\end{tabularx}
\end{table}

As summarized in Table~\ref{tab:aug_summary}, data augmentation remains a low-cost yet indispensable strategy for DI. The choice of specific technique is highly dependent on the target task and data characteristics. A common best practice is to employ basic visual transformations as a baseline and then strategically layer on advanced methods based on the primary bottleneck: physical degradation simulation for scanned documents, structural perturbations for layout-sensitive tasks, and semantic expansion for improving instruction-following capabilities. When maximum performance is required, automated augmentation offers a powerful tool for discovering the optimal policy.

Looking forward, we anticipate data augmentation will evolve toward more \textbf{automated}, \textbf{learnable}, and \textbf{controllable} paradigms. Treating augmentation as an optimizable module rather than a fixed pre-processing step, as seen in policy search, will likely become dominant. Furthermore, advances in large generative models will enable controllable augmentation, where effects can be generated from high-level instructions (e.g., ``make this invoice look rained on''), significantly enhancing flexibility.

\section{Data Generation from Scratch}
\label{sec:generation_from_scratch}
\begin{figure*}[t]
  \centering
  \includegraphics[width=\textwidth]{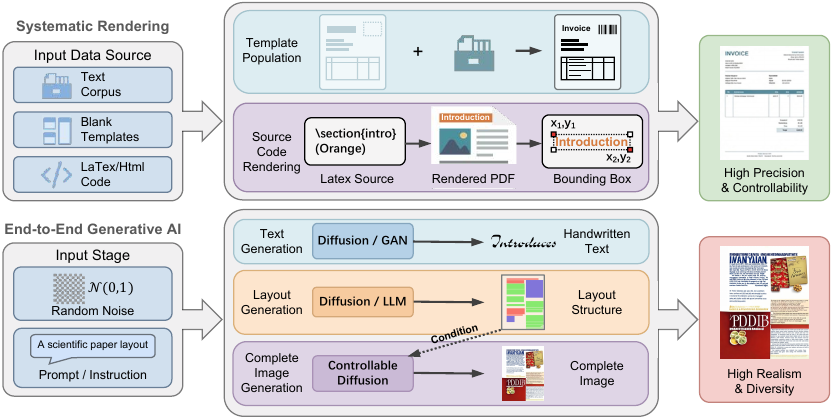}
  \caption{Comparison of Data Generation from Scratch paradigms: (1) Systematic Rendering: Populates templates with content to render documents with high annotation precision and controllability; (2) End-to-End Generative AI: Leverages diffusion models or LLMs to synthesize diverse and realistic document images, layouts, and handwritten text directly from noise or prompts, aiming to bridge the fidelity gap with real-world data.}
  \label{fig:Figure6}
\end{figure*}

The technical paradigms for data generation from scratch fall into two primary categories, as illustrated in Figure~\ref{fig:Figure6}. Section~\ref{sec:template_rendering} reviews \textbf{systematic methods based on templates, rules, and rendering}, which construct documents through deterministic pipelines. Section~\ref{sec:generative_ai} then explores \textbf{end-to-end generative methods}, which leverage modern generative AI to learn and sample from the latent distribution of real data.

\subsection{Systematic Generation via Templates, Rules, and Rendering}
\label{sec:template_rendering}
Methods based on templates and rendering operate on a distinct pipeline: a template defines the structure, a program populates the content, and a rendering engine produces the final visual output. This process generates a multimodal data package containing both the visual appearance and perfectly aligned annotations. Due to their unparalleled annotation precision and controllability, these methods serve as the cornerstone for constructing large-scale synthetic datasets.

\subsubsection{Generating Basic Components}
Generating large-scale, high-quality data of foundational components—such as text lines, tables, charts, and formulas—is a critical prerequisite for training specialized document models.

For \textbf{text lines}, a common approach is to render large text corpora (e.g., from Wikipedia) with numerous fonts to synthesize vast datasets of printed or handwritten text lines~\cite{li2023trocr, lv2023kosmos, pippi2023handwritten, nikolaidou2024rethinking}. Another method creates long, visually coherent handwritten texts by concatenating stylistically similar short samples~\cite{timofeev2023dss}.

For highly structured components like \textbf{tables}, \textbf{charts}, and \textbf{formulas}, template-based rendering is particularly effective. 
For tables, the core idea is to define a structure, populate it with content (which can be generated by LLMs or extracted from real documents), and then render the final output using engines like HTML/CSS or Pandoc~\cite{davis2022end, quattrini2024mu, bradley2024synfintabs, kim2024tablevqa}. 
For charts, the typical pipeline involves generating numerical data, rendering it into a chart image using plotting libraries (e.g., Matplotlib), and then using templates to synchronously create corresponding question-answer pairs~\cite{kahou2017figureqa, kafle2018dvqa, methani2020plotqa}. This foundational approach has been widely extended to support more diverse chart types and styles, often leveraging LLMs to generate more natural questions~\cite{chaudhry2020leaf, luo2021chartocr, liu2023matcha, xu2023chartbench, tang2023vistext, xia2023structchart, xia2025chartx, chen2024onechart}.
For mathematical formulas, generators have been built to render content from \texttt{.tex} source files into images with corresponding Markdown annotations~\cite{wei2024general}, or to synthesize handwritten formulas by rendering real symbol strokes into \LaTeX{} layouts~\cite{gervais2025mathwriting}.

\subsubsection{Generating Complete Documents}
Beyond basic components, template-based methods are also widely used to create complete, multi-element documents, with strategies varying by document type.

For \textbf{semi-structured documents} with relatively fixed layouts, a common and effective workflow is to print text onto a blank template. This approach has been widely used to generate large volumes of synthetic invoices, train tickets, and ID cards~\cite{guo2019eaten, monsur2023synthnid, wojcik2023nbid, arslan2024deep}. A more flexible strategy involves abstracting layouts into graph structures and then randomly combining structural templates with text dictionaries extracted from real documents~\cite{Wojcik2025NewPI}.

Second, for more complex \textbf{structured documents} like scientific papers and web pages, their intrinsic source code (e.g., \LaTeX{} and HTML) is often leveraged. For scientific papers, DLAgen~\cite{ying2024fine} generates pixel-level masks via a coloring scheme, where unique colors assigned at the \LaTeX{} source code are mapped back from the rendered output to create zero-error annotations. Other approaches use a model to first generate a layout before populating it with content~\cite{pisaneschi2023automatic}. For web pages, their inherent HTML structure allows for direct rendering into visual documents with perfectly aligned hierarchical annotations~\cite{kim2023web, zhang2023internlm, zhang2024internlm}.

Finally, a more general paradigm is to generate \textbf{universal documents} by assigning layout and visual properties to plain text content. This approach has been used to convert unstructured text from sources like Wikipedia into document images~\cite{yang2017learning, li2020cross, davis2022end}, and more recently, to leverage massive plain-text QA datasets (e.g., SQuAD) for training multimodal models~\cite{powalski2021going, lee2023pix2struct, aggarwal2023dublin, kim2024dockd}. To systemize this process, powerful synthesizers like SynthDoG~\cite{kim2021donut} and its variants~\cite{mao2024visually, okamoto2024crepe} have been developed, which convert arbitrary text into realistic document images using dynamic layout rules and rich rendering parameters. Other methods include pipelines for synthesizing complete bilingual documents~\cite{ding2024synthdoc}, probabilistic models for automatic layout and content generation~\cite{raman2022synthetic}, and complex rule-based systems for creating diverse business documents~\cite{vsimsa2023docile}. A notable application is HW-SQUAD~\cite{mathew2021asking}, which converted the SQuAD dataset into a large-scale handwritten document VQA dataset via font rendering and degradation effects.

\subsection{End-to-End Methods Based on Generative AI}
\label{sec:generative_ai}
A more cutting-edge paradigm uses generative AI models to directly learn the latent distribution of real document data. Unlike template-based methods, these approaches generate a document's visual appearance, layout, and content in an end-to-end manner from random noise or a conditional input, demonstrating immense potential for diversity, realism, and stylization. This section categorizes these methods by their generation target.

\subsubsection{Handwritten Text Generation}
As a challenging subfield, this task aims to synthesize text images that preserve specific content while rendering realistic handwriting styles.

\textbf{GAN-based methods} were the early mainstream. A series of works progressively improved the core GAN architecture for quality and diversity~\cite{kang2020ganwriting, fogel2020scrabblegan, mattick2021smartpatch, gan2022higan+}, sometimes combining it with Transformers to better capture sequential dependencies~\cite{davis2020text, wang2022approach}. GANs have also been used to tackle specific challenges, such as generating low-resource Tibetan characters~\cite{guo2023tibetan}, or performing cross-lingual style transfer~\cite{li2023cross}.

More recently, \textbf{diffusion-based methods} have surpassed GANs in generation quality and controllability, enabling high-quality text synthesis in challenging few- or zero-shot settings~\cite{luhman2020diffusion, nikolaidou2023wordstylist, dai2024one}. Other works have expanded their application boundaries, extending the generation scope from words to full paragraphs~\cite{mayr2025zero, pippi2025zero} or applying them to generate complex characters like Chinese~\cite{ding2023improving, gui2023zero}.

Beyond these mainstream approaches, researchers have also explored other technical paradigms, including recurrent Variational Autoencoders (VAEs) for online handwriting~\cite{aksan2018deepwriting}, RNNs for learning disentangled style descriptors~\cite{kotani2020generating}, and pure Transformer architectures for generation and style disentanglement~\cite{bhunia2021handwriting, dai2023disentangling}. Hybrid models, such as combining a VAE with a Transformer, have also been investigated~\cite{vanherle2024vatr++}.

\subsubsection{Document Layout Generation}
This task focuses on synthesizing the abstract structure of a document—defined by element categories, positions, and sizes—without rendering pixel-level content.

Early explorations predominantly employed \textbf{GANs, VAEs, or autoregressive Transformers} to generate layouts from scratch, typically modeling the layout as a discrete sequence of elements~\cite{gupta2021layouttransformer, arroyo2021variational, patil2020read}. Subsequent research aimed to solve specific challenges, such as constrained generation via latent optimization~\cite{kikuchi2021constrained}, enhancing diversity with multi-choice learning mechanisms~\cite{nguyen2021diverse}, improving structural rationality through cooperative generation~\cite{weng2023learn}, or fixing element overlaps using an innovative locator-corrector framework~\cite{lin2024spot}.

In recent years, \textbf{diffusion models} have emerged as the dominant paradigm, outperforming autoregressive models by generating layouts through an iterative denoising process that ensures higher structural coherence and diversity~\cite{kong2022blt, chai2023layoutdm, inoue2023layoutdm, zhang2023layoutdiffusion}. Subsequent innovations have further enhanced the quality and controllability by introducing flexible conditioning or simultaneously handling discrete categories and continuous coordinates~\cite{guerreiro2024layoutflow, wang2024dolfin, hui2023unifying, he2023diffusion}.

With the rise of \textbf{LLMs}, a new paradigm with a greater capacity for semantic understanding has emerged. This has evolved from leveraging in-context learning with prompts to generate layout sequences without fine-tuning~\cite{lin2023layoutprompter, jiang2023layoutformer++}, to understanding complex natural language instructions~\cite{lu2025uni}. A disruptive idea is to redefine layout generation as a code generation task, prompting an LLM to directly generate HTML/CSS code instead of coordinate sequences~\cite{tang2024layoutnuwa}. This paradigm has demonstrated powerful end-to-end capabilities, such as generating a complete poster layout from only a scientific abstract~\cite{wang2024scipostlayout}.

\subsubsection{Complete Document Image Synthesis}
Beyond isolated components, this task aims to synthesize holistic document images that simultaneously integrate coherent content, layout, and realistic visual appearance.

Early explorations relied on \textbf{autoregressive models}. For instance, DocSynth and its successor~\cite{biswas2021docsynth, biswas2024docsynthv2} represented a document as a unified sequence of categories, positions, and text, and then trained a Transformer to generate these elements sequentially for rendering.

\textbf{VAEs} have shown unique advantages for specific formats. CanvasVAE~\cite{yamaguchi2021canvasvae}, for example, generates vector graphic documents (e.g., SVG, PDF) directly from scratch. By learning and sampling from the latent space of real vector documents, it creates new, infinitely scalable and editable documents composed of structured elements.

Recently, \textbf{controllable diffusion models} have revolutionized this field by enabling precise visual synthesis. A prime example is DIG~\cite{ying2024dig}, which leverages ControlNet to condition Stable Diffusion on layout inputs, synthesizing high-fidelity documents with coherent text and styles to bridge the gap between structural control and photorealism.

\begin{table}[t]
\caption{Summary of Data Generation from Scratch Paradigms.}
\label{tab:gen_scratch_summary}
\footnotesize
\renewcommand{\arraystretch}{1.3}
\begin{tabularx}{\textwidth}{
  >{\raggedright\arraybackslash}p{3.5cm}
  >{\raggedright\arraybackslash}p{3.8cm}
  >{\raggedright\arraybackslash}X
  >{\raggedright\arraybackslash}X
}
\hline
\textbf{Generation Targets} & \textbf{Advantages} & \textbf{Limitations} & \textbf{Typical Use Cases} \\
\hline
\multicolumn{4}{l}{\textbf{1. Systematic Generation via Templates, Rules, and Rendering}} \\
\hline
Basic components: text lines, tables, charts; Complete docs: invoices, papers, web pages
& Highly controllable; \newline perfect annotations; \newline high generation efficiency; \newline strong logical consistency.
& Limited diversity; \newline realism can be challenging; \newline rule design is complex.
& Building large-scale, structured, task-oriented datasets (e.g., for OCR, Layout analysis, KIE). \\
\hline
\multicolumn{4}{l}{\textbf{2. End-to-End Methods Based on Generative AI}} \\
\hline
Handwritten text; document layouts; complete document images
& High diversity; \newline high realism in style and texture; \newline no manual rule design needed.
& Limited controllability; \newline prone to hallucinations; \newline high training cost.
& Stylized data generation (e.g., handwriting); bridging the fidelity gap in realism. \\
\hline
\end{tabularx}
\end{table}

\subsection{Observations}

Table~\ref{tab:gen_scratch_summary} compares the key paradigms discussed in this section. Fundamentally, the choice between systematic rendering and generative AI involves a trade-off between controllability/precision versus realism/diversity. Despite significant progress, the field still faces formidable challenges in generating data indistinguishable from real-world samples. These include bridging the \textbf{Fidelity Gap} in visual and content realism, ensuring \textbf{Consistency} in long-form and multimodal documents, and mitigating \textbf{Factuality and Hallucination} issues in LLM-based content generation.

Looking forward, we anticipate this field will evolve toward providing higher-quality and more complex training signals. Key future directions include: (1) \textbf{Controllable Generation}, which will enable the creation of bespoke datasets for fine-grained scientific study (e.g., systematically varying a single attribute like font size); (2) the deep integration of generative models with \textbf{Physics-Based Rendering} engines to fundamentally address the domain generalization problem; and (3) a grander vision of \textbf{World Models and Document Simulation}, which will generate collections of interconnected documents from simulated real-world processes, providing crucial contextualized data for training models on cross-document reasoning tasks.

\section{AUTOMATED DATA ANNOTATION}
\label{sec:auto_annotation}
\begin{figure*}[t]
  \centering
  \includegraphics[width=\textwidth]{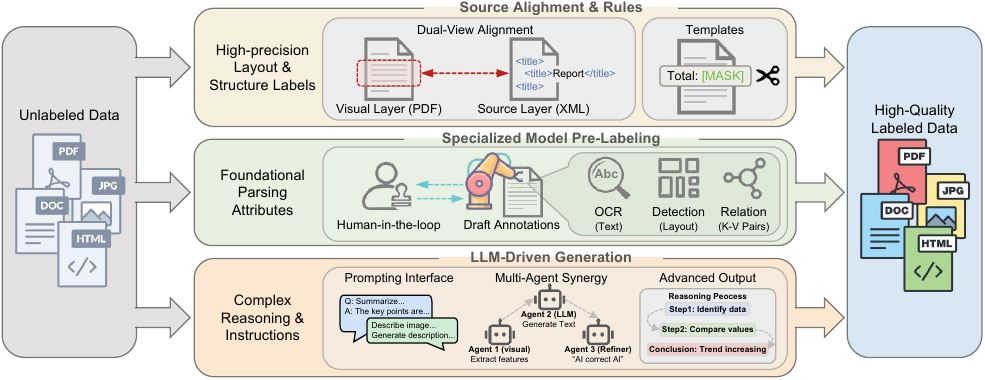}
  \caption{The evolution of Automated Data Annotation paradigms. The methods are organized into three layers based on the source of intelligence: (1) Source Alignment \& Rules, leveraging intrinsic file structures for deterministic labeling; to (2) Specialized Model Pre-Labeling, utilizing pipelined tools for draft annotations; and finally to (3) LLM-Driven Generation, employing LLMs to synthesize complex signals such as detailed descriptions, chain-of-thought rationales, and self-refined instructions.}
  \label{fig:Figure5}
\end{figure*}

The technical paradigms for automated data annotation have undergone a clear evolution. This section reviews the related work according to these core technical shifts, depicted in Figure~\ref{fig:Figure5}, which fall into three main stages. Section~\ref{sec:heuristics} introduces early methods using \textbf{external source and heuristic rules} to generate labels. Section~\ref{sec:prelabeling} delves into the current mainstream paradigm of using \textbf{automated tools and specialized models} for pre-labeling. Finally, Section~\ref{sec:llm_annotation} focuses on cutting-edge techniques driven by \textbf{LLMs}, which generate advanced annotations.

\subsection{Generating Labels from External Sources and Heuristics}
\label{sec:heuristics}
Before the widespread adoption of deep learning, researchers pioneered methods for automated annotation by leveraging external sources or designing sophisticated heuristic rules. These approaches do not rely on the predictive power of models, and their generation logic is typically highly interpretable and deterministic.

\subsubsection{Leveraging the Intrinsic Structured Representations of Documents}
Many digitally-born documents possess dual representations: a visual layer for human reading and a machine-readable source layer (e.g., HTML, XML, \LaTeX{}). By aligning these two layers, high-quality annotations can be generated for the visual document at virtually zero cost.

This principle applies across diverse source formats. For highly structured \textbf{HTML/XML}, datasets like PubLayNet~\cite{zhong2019publaynet} align semantic tags with PDF visual regions via string matching. For semi-structured \textbf{Office and \LaTeX{} documents}, researchers employ sophisticated parsing techniques—such as rendering source elements with unique identifiers (e.g., color codes) to robustly map structural logic to visual bounding boxes~\cite{weber2023wordscape, wang2021layoutreader}. Similarly, others directly parse \LaTeX{} source files to generate annotations for figures, theorems, or other logical relations~\cite{siegel2018extracting, xia2024docgenome, mishra2024modular}. For \textbf{tables}, this alignment strategy moves beyond simple bounding boxes to capture intricate cell structures. By parsing source code from HTML/XML~\cite{zhong2020image, zheng2021global}, \LaTeX{}\cite{chi2019complicated, qasim2019rethinking}, or Word documents\cite{li2020tablebank}, researchers have successfully constructed large-scale datasets with high-precision logical annotations at zero manual cost.

\subsubsection{Leveraging Layout Patterns and Task Logic}
Even without a parallel structured source, annotations can be created by leveraging a document's intrinsic layout patterns and the task's logical structure.

One approach is based on \textbf{layout and geometric rules}. Early works used classic image processing algorithms like contour detection to extract basic layout elements~\cite{shen2020large}, or skeletonization to construct training triplets from handwritten images~\cite{guan2020improving}. Later methods focused on inherent layout conventions, using heuristics like paragraph indentations to classify text lines~\cite{ma2023hrdoc}, or relative cell positions to infer table structures~\cite{yang2023large}. TreeForm~\cite{zmigrod2024treeform} further used transformation rules to convert existing flat annotations into richer, tree-structured labels that capture full table structures and page-level hierarchies.

Another major approach relies on \textbf{task logic and templates}, particularly for question-answering. The foundational paradigm is template-based QA synthesis, where questions are generated by programmatically slotting structured document information into predefined linguistic patterns. This strategy has been widely applied to complex layouts, generating sophisticated questions for book covers~\cite{mishra2019ocr}, software specifications~\cite{dhingra2017quasar}, cooking recipes~\cite{yagcioglu2018recipeqa}, and tabular data~\cite{raja2023icdar}. Most recently, this strategy has matured into a paradigm of annotation format conversion. Instead of generating QA pairs from raw data, these methods repurpose existing structured annotations (e.g., KIE bounding boxes) into unified instruction-response or VQA formats, addressing the scarcity of labels for target tasks and effectively unlocking the value of legacy datasets~\cite{cha2024honeybee, zmigrod2024value}.

\subsection{Generating Labels with Automated Tools and Specialized Models}
\label{sec:prelabeling}
With the rise of deep learning, using automated tools or pre-trained specialized models to generate draft annotations for unlabeled documents has become a mainstream strategy. This section reviews these methods, categorized by the type of annotations they generate.

\subsubsection{Generating Foundational Parsing Annotations}
As the starting point of the pre-labeling workflow, this category focuses on extracting core document elements: textual content, physical layout, and structural attributes.

For \textbf{textual content transcription}, a vast body of research relies on off-the-shelf OCR engines to digitize document collections~\cite{yu2024texthawk2, kim2021donut, yu2023documentnet, ouyang2025omnidocbench, yang2025cc, ding2024david, ding2024mmvqa, lyu2024structextv3, liu2024focus, wei2024general}. Similarly, machine translation models are widely used to generate cross-lingual content annotations for low-resource languages~\cite{chen2024far, tahsin2021deep}.

For \textbf{physical layout and structural attributes}, methodologies target different levels of granularity.
At the macroscopic level, layout annotation strategies fall into three categories. Parsing-based methods utilize tools like PDFMiner and GROBID to directly extract coordinates and semantic labels~\cite{ding2024david, ding2024mmvqa, pisaneschi2023automatic}. Vision-based methods, in contrast, leverage pre-trained object detectors (e.g., YOLOv7, Mask-RCNN) to generate weak labels\cite{ahuja2023new, ding2023vqa}. Finally, multimodal pipelines integrate powerful models like LayoutLMv3\cite{huang2022layoutlmv3} and UniMERNet\cite{wang2024unimernet} to produce comprehensive annotations for layouts, formulas, and tables\cite{ouyang2025omnidocbench, yang2025cc}. At the granular level, methods focus on finer structural attributes, such as predicting page numbers from source code~\cite{blecher2024nougat}, generating character-level segmentation points for online handwriting~\cite{kotani2020generating, aksan2018deepwriting}, or parsing PDF metadata to create pixel-level render layer masks (e.g., text, vector, raster)~\cite{li2020cross}.

\subsubsection{Generating High-Level Semantic Annotations}
Building on foundational parsing, specialized models are used to generate more complex, semantically rich annotations like entity relations and simple question-answer pairs.

For \textbf{entity and relation annotation}, methodologies range from model-based inference to knowledge-driven matching. Some works leverage pre-trained models like LayoutLMv3 to generate generic Key-Value pairs~\cite{shukla2022dosa}, or use text classifiers to produce weak labels for entity categories~\cite{yu2023documentnet}. Others, like BiblioPage~\cite{kohut2025bibliopage}, incorporate external knowledge by matching OCR text against library catalog metadata to generate annotation proposals.

For generating \textbf{simple question-answering annotations}, pipelines typically follow either an end-to-end or a semi-automatic paradigm. ChartQA~\cite{masry2022chartqa} exemplifies the end-to-end approach, fine-tuning T5 models to sequentially generate answers and questions from summaries. In contrast, semi-automatic methods first employ traditional NLP tools (e.g., noun phrase extractors) to identify candidate answers, before using generative models to synthesize the corresponding questions~\cite{yang2017semi, saikh2020scholarlyread, saikh2022scienceqa}.

\subsection{Generating Task Annotations with LLMs}
\label{sec:llm_annotation}
The advent of LLMs has fundamentally transformed automated annotation. Unlike conventional models restricted to discriminative labeling, LLMs leverage their generative capabilities to synthesize complex, task-oriented signals—ranging from multi-turn dialogues to reasoning chains. This shift enables the low-cost construction of training data for advanced cognitive tasks that were previously reliant on expert manual annotation.

\subsubsection{Generating Task-Oriented Annotations}
The primary utility of LLMs lies in synthesizing supervisory signals for downstream tasks. Methodologies in this category fall into two streams: interactive query-response generation and descriptive content synthesis.

In the realm of \textbf{question-answering and dialogue annotation}, research has progressed from simple QA generation to handling complex interaction scenarios. A vast body of work leverages powerful LLMs to construct large-scale QA pairs for diverse document types~\cite{laurenccon2024building, faysse2025colpali, kim2024dockd, ding2024mmvqa, xia2024docgenome, kim2024tablevqa, liu2024mmc, lyu2024structextv3, wang2023layout}. This paradigm has been further extended to complex extensions involving multi-turn dialogues with or without grounding coordinates~\cite{zhang2023llavar, wang2023towards, feng2023unidoc}, and the synthesis of negative samples with logical fallacies to enhance robustness~\cite{liu2024mitigating, li2024multimodal}. Hybrid workflows, such as ``AI-proposes, human-answers,'' have also emerged to ensure high precision for charts and maps~\cite{li2024seed}.

For \textbf{descriptive text annotation}, approaches have evolved from single-step prompting to multi-stage refinement pipelines. Direct generation is effectively used to synthesize initial descriptions from scratch~\cite{liu2024focus, masry2023unichart, yehudai2024achieving} or to upgrade existing low-quality labels~\cite{li2025llavaonevision, luo2025mono, zhao2024harmonizing}. For scenarios demanding higher fidelity, collaborative refinement strategies are employed. For instance, MiniGPT-4~\cite{zhu2024minigpt} introduces an ``AI corrects AI'' mechanism, where one model's preliminary description is refined by another. Taking this further, Monkey~\cite{li2024monkey} utilizes specialized vision experts (e.g., BLIP2, SAM) with LLMs to produce hierarchical, detail-rich descriptions.

\begin{table}[t]
\caption{Summary of Automated Data Annotation Paradigms.}
\label{tab:auto_annotation_summary}
\footnotesize
\renewcommand{\arraystretch}{1.3}
\begin{tabularx}{\textwidth}{
  >{\raggedright\arraybackslash}p{3.5cm}
  >{\raggedright\arraybackslash}X
  >{\raggedright\arraybackslash}X
  >{\raggedright\arraybackslash}X
}
\hline
\textbf{Techniques} & \textbf{Advantages} & \textbf{Limitations} & \textbf{Typical Use Cases} \\
\hline
\multicolumn{4}{l}{\textbf{1. From External Sources \& Heuristics}} \\
\hline
XML/\LaTeX{} alignment, layout/task heuristics
& High precision, highly interpretable
& Heavily relies on source data or fixed rules; poor generalization
& Digitally-born documents; documents with fixed layouts; basic document parsing tasks \\
\hline
\multicolumn{4}{l}{\textbf{2. With Automated Tools \& Specialized Models}} \\
\hline
OCR engines, translation models, object detectors
& Significantly improves annotation efficiency; mature and widely used
& Model predictions may contain systematic biases; still requires human review effort
& Standard workflow for large-scale dataset creation; accelerating draft annotations \\
\hline
\multicolumn{4}{l}{\textbf{3. With Large Language Models (LLMs)}} \\
\hline
Instruction/QA generation, CoT/rationale generation
& Can generate complex annotations for cognitive tasks
& Heavily relies on powerful LLMs; hallucination issues; high verification cost
& Building training data for instruction fine-tuning and complex reasoning tasks \\
\hline
\end{tabularx}
\end{table}

\subsubsection{Generating Reasoning Process and Instructional Annotations}
Beyond simply producing the ``output'', LLMs excel at articulating the ``process'' and refining the ``input,'' providing explicit supervision for reasoning and instruction-following.

One category is \textbf{reasoning process annotation}, which focuses on exposing the intermediate steps of cognition. Early efforts prompted LLMs to generate textual rationales or step-by-step explanations~\cite{lee2024meteor, hu2024mplug}. More advanced methods integrate spatial context, synthesizing layout-aware chains of thought that interleave textual reasoning with bounding box evidence~\cite{luo2024layoutllm, liao2025doclayllm}. Some systems, like OmniParser V2~\cite{yu2025omniparser}, adopt a scaffolded generation strategy, first predicting intermediate structural points (e.g., text centers) before refining them into detailed annotations (e.g., polygons and content).

Another category is \textbf{instructional annotation}. Methodologies for generating training instructions have witnessed a clear evolution in complexity. Pioneering works like LLaVA~\cite{liu2023visual} demonstrated the feasibility of synthesizing instructions from scratch. Subsequent research improved diversity by using templates to rephrase seed instructions~\cite{liu2024hrvda, tanaka2024instructdoc}, and later advanced to complex rewriting, utilizing Chain-of-Thought prompting to upgrade simple commands into intricate tasks~\cite{zhang2024internlm}. Most recently, self-refining loops have emerged, exemplified by TextSquare~\cite{tang2024textsquare}, where the model iteratively questions, answers, and evaluates itself to autonomously curate high-quality instruction datasets.

\subsection{Observations}
As summarized in Table~\ref{tab:auto_annotation_summary}, the landscape of automated data annotation has expanded from deterministic programmatic alignment to generative LLM-driven synthesis. The optimal strategy is dictated by the complexity of the target annotation: rigorous programmatic methods remain the gold standard for simple, deterministic labels, while LLMs are indispensable for synthesizing high-level semantic reasoning.
Despite its efficiency, this paradigm introduces significant reliability bottlenecks, notably the high \textbf{verification cost} for complex generative outputs and the risk of \textbf{bias amplification} from teacher models. Future research must prioritize automated verification mechanisms and confidence calibration to ensure that the scale of automated annotation does not compromise its trustworthiness.

\begin{figure*}[t]
  \centering
  \includegraphics[width=0.9\textwidth]{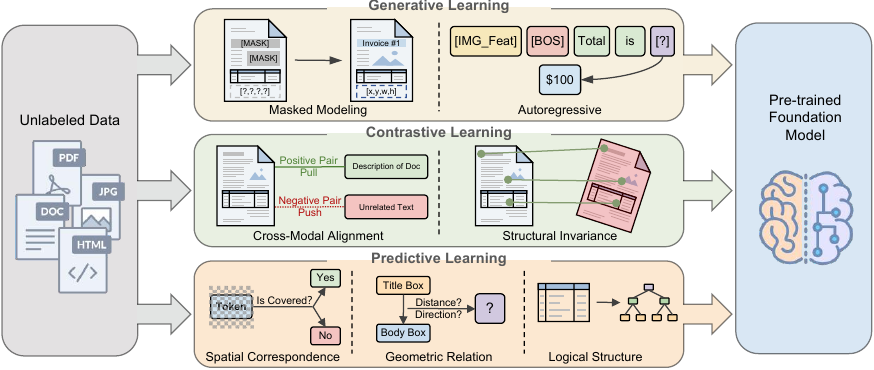}
  \caption{Taxonomy of Self-Supervised Signal Construction. The methods mines supervision from unlabeled documents through three mechanisms: (1) Generative Learning: Generative Learning: Forces the model to recover masked content or autoregressively predict the next token; (2) Contrastive Learning: Learns invariant features by pulling together positive pairs and pushing apart negative ones; and (3) Predictive Learning: Trains the model to infer intrinsic document properties.}
  \label{fig:Figure7}
\end{figure*}

\section{Self-Supervised Signal Construction}
\label{sec:self_supervised}
As summarized in Figure~\ref{fig:Figure7}, this section categorizes self-supervised methodologies into three paradigms: \textbf{Generative Learning} (Section~\ref{sec:generative_learning}), \textbf{Contrastive Learning} (Section~\ref{sec:contrastive_learning}), and \textbf{Predictive Learning} (Section~\ref{sec:predictive_learning}). These approaches are designed for the ``abundant unlabeled data'' scenario, aiming to mine supervisory signals directly from the intrinsic structures of documents.
In this context, we adopt a broad definition of ``unlabeled data'': documents that lack manual annotations for downstream tasks but possess inherent, extractable structures. Consequently, low-cost attributes obtainable at scale—such as OCR text, HTML DOM trees, or \LaTeX{} source code—serve as the foundational ``free labels'' for constructing these self-supervised pretext tasks.

\subsection{Generative Learning}
\label{sec:generative_learning}
Generative learning, a central paradigm in self-supervised learning, forces a model to learn rich contextual representations by training it to ``recover the whole from a part.'' In DI, this paradigm is primarily divided into two major categories based on the prediction objective: masked modeling and autoregressive modeling.

\subsubsection{Masked Modeling}
The core of masked modeling is a ``fill-in-the-blank'' task that leverages bidirectional context to learn deep contextual representations. This cornerstone of document foundation models has evolved from unimodal to multimodal fusion.

In the \textbf{text modality}, LayoutLM~\cite{xu2020layoutlm} pioneered the Masked Visual-Language Model (MVLM), requiring the model to predict masked tokens using both textual context and 2D layout information. This paradigm was inherited by subsequent works~\cite{xu2021layoutlmv2, xu2021layoutxlm} and validated for in-domain pre-training~\cite{vsimsa2023docile}.
In the \textbf{vision modality}, DiT~\cite{li2022dit}, inspired by BEiT~\cite{bao2022beit}, proposed Masked Image Modeling (MIM) to predict discrete visual tokens for masked document image patches. In contrast to predicting discrete tokens, DocumentNet~\cite{yu2023documentnet} required pixel-level reconstruction of masked patches to learn a finer-grained correspondence.
For the document-specific \textbf{layout modality}, researchers also designed specialized tasks. BLT~\cite{kong2022blt}, for instance, proposed Masked Attribute Prediction (MAP), which randomly masks any attribute (e.g., category, coordinates, or size) in a sequence of layout elements for the model to recover. A similar masked coordinate prediction task was also explored in~\cite{feng2023sequence}.

As the field advanced, a shift towards \textbf{joint multimodal masking} occurred. Early explorations synchronously masked corresponding text and vision regions~\cite{powalski2021going}. LayoutLMv3~\cite{huang2022layoutlmv3} advanced this unification by treating text tokens and image patches as homogeneous inputs within a single Transformer, applying masking strategies indiscriminately across modalities to enforce deep cross-modal alignment. Subsequent works pursued even more comprehensive fusion, proposing joint text-layout reconstruction tasks that require simultaneous prediction of both masked text and its layout position~\cite{tang2023unifying, luo2024layoutllm}.

\subsubsection{Autoregressive Modeling}
The core of autoregressive modeling is unidirectional sequence prediction, training a model to predict the next element based on preceding ones to learn sequence coherence. In DI, this is applied via two main technical paths.

The first path is \textbf{language modeling on a symbolic sequence}, which does not operate directly on pixels. Instead, a document's visual and textual information is pre-encoded into a unified, interleaved 1D sequence of symbols, on which the model performs next-token prediction. This is a foundational objective for modern MLLMs~\cite{zhang2023internlm, laurenccon2024building, luo2025mono, yu2024texthawk2, chen2024expanding}. A key variation is text infilling, which, unlike parallel masked modeling, first masks long text spans and then autoregressively regenerates the content word by word. This task is used to enhance contextual understanding and generative capabilities, as explored in several works~\cite{feng2023sequence, wang2024docllm, davis2022end, lee2023pix2struct}.

The second path reframes a document perception task as an holistic autoregressive \textbf{image-to-sequence generation task}, taking the 2D document image as direct input and train the model to generate a 1D sequence that describes it. The pseudo-OCR task pioneered by Donut~\cite{kim2021donut} is the most influential idea here. It trains a model to autoregressively generate all text content from an image, implicitly teaching it powerful OCR capabilities. This end-to-end image-to-text approach has been widely adopted and extended~\cite{feng2024docpedia, liu2024hrvda, wei2024general, luo2024layoutllm}. Variations include narrowing the input to a segmented region~\cite{cao2023attention} or extending the output from plain text to a mixed sequence of text and layout~\cite{biswas2024docsynthv2}.

\subsection{Contrastive Learning}
\label{sec:contrastive_learning}
Contrastive learning, another powerful self-supervised paradigm, learns invariant features by constructing positive and negative sample pairs to pull similar samples closer and push dissimilar ones apart in the feature space. Its application in DI shows a clear evolution of the contrastive objective from concrete to abstract.

Early explorations began with \textbf{local visual consistency}, where a positive pair consisted of any two pixel features within the same coarse bounding box, encouraging local feature similarity~\cite{he2017multi}. Subsequently, the objective extended to \textbf{cross-modal alignment}, evolving from a macroscopic to a microscopic scale. At the macroscopic level, the Text-Image Matching (TIM) task in LayoutLMv2~\cite{xu2021layoutlmv2} treats the entire image-text pair from the same document as a positive sample to learn document-level semantic consistency, a task later extended to multilingual scenarios by LayoutXLM~\cite{xu2021layoutxlm}. To address the feature collapse  problem  where document-level contrast can overlook fine-grained details, DoCo~\cite{li2024enhancing} proposed a more granular Document Object Contrastive learning at the microscopic level. It focuses on each individual object (e.g., a text block) and treats its pure visual features and multimodal features as a positive pair to learn object-level cross-modal representation consistency.

The most advanced applications aim to learn invariance to more \textbf{abstract, content-independent attributes}. DocReL~\cite{li2022relational} pushed the objective to the level of ``structural relationships'' by treating two differently augmented versions of the same document as a positive pair, learning a representation of structure that is invariant to visual perturbations. 
Similarly, \cite{dai2023disentangling} applied this idea to disentangle handwriting style. It achieves this by designing two parallel contrastive tasks: a writer-style task where different characters from the same author form a positive pair, and a character-style task where different stroke regions within the same character form a positive pair, thereby learning style representations that are invariant to content.

\subsection{Predictive Learning}
\label{sec:predictive_learning}
Beyond generative and contrastive learning, predictive learning trains a model to explicitly predict intrinsic, programmatically extractable structures, relations, or attributes of a document. Unlike generative learning, which ``recovers'' raw input, predictive learning teaches the model to ``infer'' higher-level, abstract information. These tasks can be grouped into three categories based on their level of abstraction.

First is the prediction of \textbf{spatial correspondence} to learn fine-grained text-image alignment. Pioneering works like LayoutLMv2~\cite{xu2021layoutlmv2} and LayoutLMv3~\cite{huang2022layoutlmv3} introduced introduced alignment tasks—specifically Text-Image Alignment (TIA) and Word-Patch Alignment (WPA)—that train the model to predict whether a text token's corresponding image region is covered or masked. Similarly, tasks in SeRum~\cite{cao2023attention} train the model to predict a segmentation mask corresponding to a given query or text segment.

Second is the prediction of \textbf{geometric attributes and relations}, which requires computing specific geometric values. This includes learning the bidirectional mapping between text content and its bounding box coordinates~\cite{aggarwal2023dublin, lu2025bounding}, or more advanced tasks such as computing the relative distance and direction between two text elements~\cite{liao2025doclayllm}.

The highest-level tasks involve predicting \textbf{logical and hierarchical relations}. For web documents, MarkupLM~\cite{li2022markuplm} introduced tasks like predicting the structural relationship (e.g., parent-child) between two nodes in a DOM tree, forcing the model to encode the hierarchical grammar of web pages. Distinct from the pseudo-OCR task in Section~\ref{sec:generative_learning} that transcribes visible text, a more challenging image-to-Markdown task requires the model to infer the implicit logical structure of the image and generate Markdown text that preserves it. This was proposed by \cite{lv2023kosmos} and \cite{hu2024mplug}, with the latter unifying the idea across documents, tables, and charts.

\subsection{Observations}

\begin{table}[t]
\caption{Summary of Self-Supervised Signal Construction Paradigms.}
\label{tab:ssl_summary}
\footnotesize
\renewcommand{\arraystretch}{1.3}
\begin{tabularx}{\textwidth}{
  >{\raggedright\arraybackslash}p{3.5cm}
  >{\raggedright\arraybackslash}p{3.5cm}
  >{\raggedright\arraybackslash}p{3.5cm}
  >{\raggedright\arraybackslash}X
}
\hline
\textbf{Pretext Tasks} & \textbf{Advantages} & \textbf{Limitations} & \textbf{Typical Use Cases} \\
\hline
\multicolumn{4}{l}{\textbf{1. Generative Learning: Recovering the whole from a part}} \\
\hline
Masked Modeling; \newline Autoregressive Modeling
& Rich supervisory signals from every token/patch; \newline strong generative capabilities.
& `[MASK]` token creates pre-train-finetune gap.
& Pre-training general-purpose, context-aware document representations. \\
\hline
\multicolumn{4}{l}{\textbf{2. Contrastive Learning: Learning invariance from sample pairs}} \\
\hline
Cross-modal consistency; \newline Instance discrimination; \newline Style disentanglement
& Excels at learning high-level abstract representations; \newline robust to data augmentation; \newline effective for disentanglement.
& Relies on high-quality positive/negative pair construction; \newline risk of mode collapse.
& Learning multimodal alignment; document retrieval; style-related tasks; relationship modeling. \\
\hline
\multicolumn{4}{l}{\textbf{3. Predictive Learning: Predicting intrinsic document properties}} \\
\hline
Geometric property prediction; \newline Logical relation prediction; \newline Structured text generation
& Injects explicit structured prior knowledge into the model; \newline highly interpretable.
& Depends on low-cost labels (e.g., from OCR, DOM trees); \newline task design is domain-specific.
& Training models with strong requirements for geometric, layout, or logical structures. \\
\hline
\end{tabularx}
\end{table}

Table~\ref{tab:ssl_summary} summarizes the three pillars of self-supervised signal construction. The design of pretext tasks is essentially a balancing act: generative learning excels at learning deep contextual representations, contrastive learning is powerful for learning invariant features, and predictive learning is tailored for teaching models specific intrinsic document properties. Modern foundation models often combine multiple pretext tasks to leverage their complementary strengths. However, the field still faces significant challenges, including mitigating the \textbf{Pre-train-Finetune Gap}, reducing the immense \textbf{computational cost}, and designing more effective, \textbf{multimodally-aligned pretext tasks} that can integrate world knowledge. Addressing these challenges is key to propelling document foundation models toward a future of higher performance and broader applicability.

\section{Evaluation of Generated Data}
\label{sec:evaluation}
A fundamental question persists throughout the data generation lifecycle: how do we know if the generated data is good? To systematically answer this, this section establishes the first comprehensive, multi-level evaluation framework for data generation in DI, comprising two fundamental paradigms. \textbf{Intrinsic Evaluation} aims to directly measure the quality of the data (e.g., fidelity, diversity) without reliance on downstream tasks. In contrast, \textbf{Extrinsic Evaluation} measures the data's effectiveness based on downstream model performance. Intrinsic evaluation ensures the data is high-quality, while extrinsic evaluation verifies it is useful. Accordingly, Section~\ref{sec:intrinsic_eval} categorizes intrinsic metrics into macro (dataset-level) and micro (sample-level) tiers. Section~\ref{sec:extrinsic_eval} examines the quantitative performance gains on downstream benchmarks. Finally, Section~\ref{sec:eval_challenges} explores the open challenges and future directions in data evaluation.

\subsection{Intrinsic Evaluation}
\label{sec:intrinsic_eval}
\subsubsection{Macro-level Quality Assessment}
Macro-level assessment evaluates the holistic quality of a generated dataset, focusing on \textbf{fidelity}, \textbf{diversity}, and \textbf{distributional alignment} with real data.

For \textbf{general distribution assessment}, the field has widely adopted standard generative metrics. The Inception Score (IS) serves as a baseline for intrinsic quality, while the Fr\'echet Inception Distance (FID) and Kernel Inception Distance (KID) measure the distributional distance between real and synthetic data in feature space. To address FID's limitations in distinguishing fidelity from diversity, disentangled metrics like Precision and Recall\cite{kynkaanniemi2019improved} and their robust variants, Density and Coverage\cite{naeem2020reliable}, are increasingly employed to diagnose specific failure modes such as mode collapse.

Beyond these general-purpose metrics, researchers have developed \textbf{targeted strategies for specific document attributes}. Diversity assessment often employs Type-Token Ratio (TTR) and Self-BLEU to quantify content variety at the vocabulary and sentence levels, or classifiers like HTG\_style\cite{nikolaidou2024rethinking} to measure handwriting style coverage. For layout distribution, READ\cite{patil2020read} introduced DocSim, a structural similarity metric that computes the average nearest-neighbor distance between real and generated layouts, explicitly accounting for semantic class and spatial geometry.

\subsubsection{Micro-level Quality Assessment}
Micro-level assessment verifies the correctness, rationality, and consistency of individual samples. We categorize methodologies into three evolutionary stages.

The first class comprises \textbf{computation-based objective metrics}, providing deterministic scores based on mathematical formulations. For visual fidelity, standard metrics like PSNR, SSIM, and LPIPS are commonly employed to measure distortion and perceptual similarity. For semantic quality, Perplexity, BLEU, and BERTScore remain the default choices. More critical to DI are structure-aware metrics: CLIP Score~\cite{hessel2021clipscore} quantifies image-text alignment, while Doc-EMD~\cite{he2023diffusion} and LTSim~\cite{otani2024ltsim} calculate the transformation cost between layouts. For hierarchical structures, Tree-Edit-Distance (TEDS) evaluates the topological correctness of HTML/\LaTeX{} trees.

The second class, \textbf{learning-based perceptual evaluators}, mimics human judgment on specific dimensions, such as predicting visual readability scores~\cite{li2019towards}, filtering low-confidence handwritten samples~\cite{ding2023improving} or using MRC models to discard simple, non-inferential data~\cite{zhang2018record}. Notably, training these evaluators often relies on synthesized data with quality labels, revealing an important trend: high-quality data generation is becoming a key enabler for building next-generation evaluation systems. Advanced systems like Uni-Layout~\cite{lu2025uni} even train evaluators on large-scale human preference datasets to judge layout aesthetics, bridging the gap between objective metrics and subjective perception.

The third and most recent class is \textbf{leveraging LLMs as evaluators}. Utilizing their general reasoning capabilities, LLMs assess semantic attributes: scoring the factuality and consistency of text~\cite{fu2023chain, gu2023don}, verifying the faithfulness of summaries against source documents~\cite{yehudai2024achieving}, and grading instructions on solvability, clarity, and hallucinations~\cite{zhang2025oasis, lee2024meteor}. This evaluation can be integrated into closed-loop systems where the model iteratively refines its own output~\cite{tang2024textsquare}. Beyond content quality, LLMs validate sample utility by detecting duplicates~\cite{chen2024expanding}, identifying modality redundancy in VQA~\cite{liu2024mmbench}, or generating decision-support labels in human-in-the-loop workflows~\cite{ma2024mmlongbench}. Finally, LLMs are used to quantify intrinsic complexity, such as tagging instruction steps~\cite{shen2025proctag} or assigning real-time error scores~\cite{iwai2024layout, lin2024spot}.

\subsection{Extrinsic Evaluation}
\label{sec:extrinsic_eval}
\subsubsection{Core Methodology}
The most direct method of extrinsic evaluation is to measure the contribution of generated data to downstream task performance. The standard workflow involves training a model utilizing generated supervisory signals—whether in the form of synthetic samples, augmented views, or self-supervised pretext tasks—and evaluating it on a real, independent test set. Consequently, the primary metric is the performance increment achieved over a baseline model trained without these generated signals. To ensure a reliable evaluation, the test set must be decoupled from the training set, for instance, by using resampling strategies to avoid template overlap~\cite{laatiri2023information}.
Beyond in-distribution accuracy, a more rigorous test measures the data's contribution to out-of-distribution (OOD) robustness. This can be achieved by constructing test sets with different layout distributions~\cite{chen2023layout} or by using comprehensive benchmarks with real-world document perturbations, such as RoDLA~\cite{chen2024rodla}, which proposes the mean Robustness Drop (mRD) as a key metric.

Apart from performance evaluation, a deeper analysis includes data efficiency and scaling effects. Data efficiency assesses ``how much synthetic data is equivalent to how much real data,'' with some high-quality generated datasets demonstrating immense value by achieving the same performance with significantly less synthetic data~\cite{shen2025proctag}. Scaling effects are studied by plotting a ``performance vs. data scale'' curve to determine whether a generation method follows a sustainable scaling law or suffers from early saturation.

\subsubsection{Task-Paradigm Mapping and Utility Analysis}
\begin{figure*}[t]
  \centering
  \includegraphics[width=\textwidth]{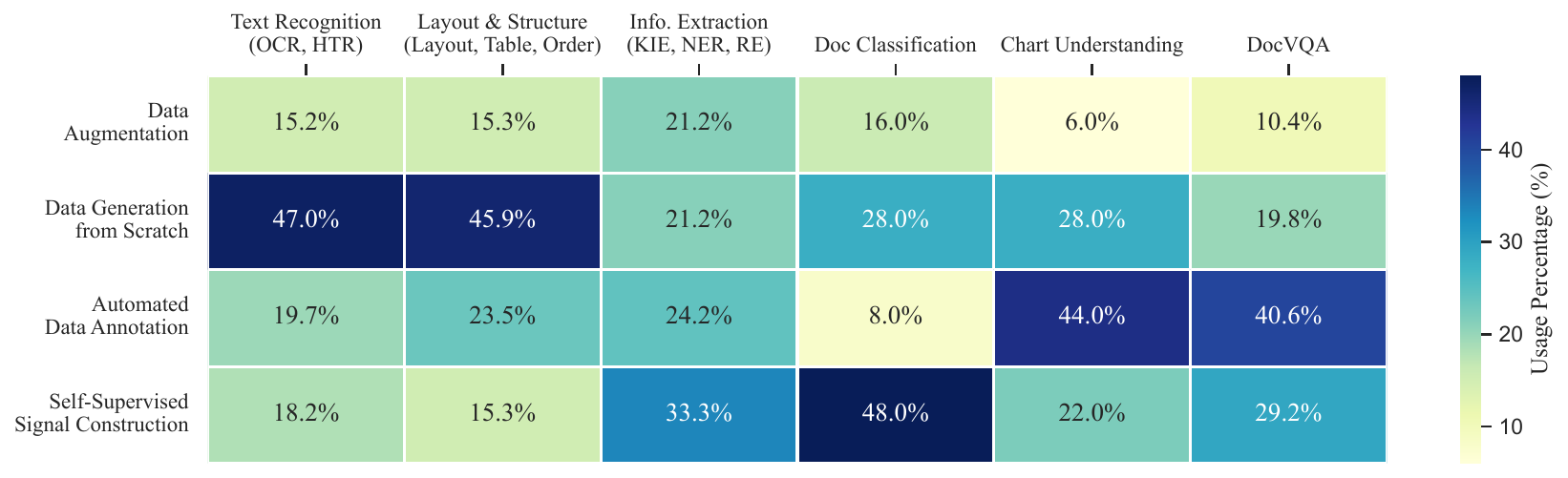}
  \caption{The Task-Paradigm Mapping Heatmap. It displays the percentage of surveyed works adopting each data generation paradigm for specific downstream tasks. Darker colors indicate a higher prevalence of the paradigm in addressing the corresponding task.}
  \label{fig:heatmap}
\end{figure*}

To synthesize the relationship between the proposed paradigms and downstream applications, we conducted a systematic statistical analysis of the surveyed literature. Figure~\ref{fig:heatmap} visualizes the distribution of research efforts, mapping each data generation paradigm to specific downstream tasks. The heatmap reveals a clear ``Perception-Reasoning Shift'' in technology selection. Specifically, Data Generation from Scratch dominates foundational perception tasks (accounting for 47.0\% of Text Recognition and 45.9\% of Layout \& Structure research), reflecting the necessity of synthesizing pixel-perfect ground truth. In contrast, for high-level reasoning tasks, the field shifts towards Automated Data Annotation, which leads in Chart Understanding (44.0\%) and DocVQA (40.6\%), driven by the capability of LLMs to generate complex QA pairs. Notably, Self-Supervised Signal Construction exhibits a broad, high-impact distribution across the board, highlighting its role as a universal semantic infrastructure.

To provide granular evidence supporting these trends, we compile the quantitative performance gains of representative methods directly in this section. To ensure analytical rigor, we only include results where the performance increment is explicitly demonstrated through ablation studies or baseline comparisons and is attributable solely to the data generation strategy. Due to variations in experimental setups across papers, the focus here is not on absolute performance, but on the performance increment achieved within each respective study. 
In the following analysis, we organize these benchmarks into six clusters aligned with the ``Perception-Reasoning'' spectrum in Figure~\ref{fig:heatmap}. For each cluster, we first identify the specific data bottlenecks (e.g., scarcity of handwriting, lack of logical labels) and then dissect how different generation paradigms effectively address these challenges to yield performance gains.

\begin{table}[t]
\caption{Performance Gains on \textbf{Document Parsing} Tasks.}
\label{tab:Document Parsing}
\scriptsize 
\renewcommand{\arraystretch}{1.3}
\begin{tabularx}{\textwidth}{
  c 
  >{\raggedright\arraybackslash}p{1.6cm}
  c 
  >{\raggedright\arraybackslash}p{1.5cm}
  >{\raggedright\arraybackslash}p{1.1cm}
  >{\raggedright\arraybackslash}p{1.3cm}
  >{\raggedright\arraybackslash}X 
}
\toprule
\textbf{Task} & \textbf{Key Work} & \textbf{Type} & \textbf{Benchmark} & \textbf{Metric} & \textbf{Gain} & \textbf{Key Contribution Highlight} \\
\midrule

\multirow{19}{*}{\rotatebox{90}{\textbf{\textit{OCR / HTR}}}} 

& DSS~\cite{timofeev2023dss}
& \gdot{coloraug}
& IAMonDo
& CER ($\downarrow$)
& 7.07 $\rightarrow$ \textbf{4.42}
& Concatenating stylistically similar handwriting samples. \\

& FgAA~\cite{chen2025fine}
& \gdot{coloraug}
& IAM
& \begin{tabular}[t]{@{}l@{}}WER ($\downarrow$) \\ CER ($\downarrow$) \end{tabular}
& \begin{tabular}[t]{@{}l@{}}15.36 $\rightarrow$ \textbf{11.10} \\ 5.51 $\rightarrow$ \textbf{3.97} \end{tabular}
& Applying fine-grained, stroke-level data augmentations. \\

& TrOCR~\cite{li2023trocr}
& \gdot{coloraug}\gdot{colorgan}
& SROIE
& F1 ($\uparrow$)
& 72.75 $\rightarrow$ \textbf{95.84}
& Pre-training solely on synthetic text lines generated by rendering engines. \\

& ScrabbleGAN~\cite{fogel2020scrabblegan}
& \gdot{colorgan}
& \begin{tabular}[t]{@{}l@{}}RIMES \\ IAM \end{tabular}
& \begin{tabular}[t]{@{}l@{}}WER ($\downarrow$) \\ NED ($\downarrow$) \end{tabular}
& \begin{tabular}[t]{@{}l@{}}12.29 $\rightarrow$ \textbf{11.68} \\ 25.10 $\rightarrow$ \textbf{23.98} \end{tabular}
& Generating varying-length handwritten text with a semi-supervised GAN. \\

& SmartPatch~\cite{mattick2021smartpatch}
& \gdot{colorgan}
& IAM
& CER ($\downarrow$)
& 4.68 $\rightarrow$ \textbf{4.04}
& Introducing a local patch-based discriminator into GAN. \\

& WordStylist~\cite{nikolaidou2023wordstylist}
& \gdot{colorgan}
& IAM
& \begin{tabular}[t]{@{}l@{}}WER ($\downarrow$) \\ CER ($\downarrow$) \end{tabular}
& \begin{tabular}[t]{@{}l@{}}14.11 $\rightarrow$ \textbf{13.28} \\ 4.86 $\rightarrow$ \textbf{4.67} \end{tabular}
& Generating stylized handwritten words using a diffusion model. \\

& VATr~\cite{pippi2023handwritten}
& \gdot{colorgan}
& \begin{tabular}[t]{@{}l@{}}Washington \\ Saint Gall \end{tabular}
& WER ($\downarrow$)
& \begin{tabular}[t]{@{}l@{}}15.9 $\rightarrow$ \textbf{13.1} \\ 32.5 $\rightarrow$ \textbf{30.9} \end{tabular}
& Generating handwritten images from visual archetypes. \\

& VATr++~\cite{vanherle2024vatr++}
& \gdot{colorgan}  
& Balanced set
& CER ($\downarrow$)
& 13.07 $\rightarrow$ \textbf{6.54} 
& Balancing character distribution via text augmentation during generation. \\

& MathWriting~\cite{gervais2025mathwriting}
& \gdot{colorgan}
& MathWriting
& CER ($\downarrow$)
& 6.2 $\rightarrow$ \textbf{5.49}
& Rendering real symbol strokes into \LaTeX{} layouts. \\

& \cite{mayr2025zero}
& \gdot{colorgan}
& CVL
& CER ($\downarrow$)
& 16.36 $\rightarrow$ \textbf{12.73}
& Generating handwritten paragraph conditioned on style images. \\

& GC-DDPM~\cite{ding2023improving}
& \gdot{colorgan}\gdot{coloranno}
& IAM
& \begin{tabular}[t]{@{}l@{}}WER ($\downarrow$) \\ CER ($\downarrow$) \end{tabular}
& \begin{tabular}[t]{@{}l@{}}19.47 $\rightarrow$ \textbf{11.57} \\ 7.27 $\rightarrow$ \textbf{3.88} \end{tabular}
& Filtering generated data using a progressive, confidence-based strategy. \\

& UniDoc~\cite{feng2023unidoc}
& \gdot{coloranno}\gdot{colorssl}
& CTW1500
& F-score ($\uparrow$)
& 0.00 $\rightarrow$ \textbf{38.27}
& Generating instruction by creating dialogues and OCR tasks. \\

& OmniParser V2~\cite{yu2025omniparser}
& \gdot{colorssl}
& \begin{tabular}[t]{@{}l@{}}Total-Text \\ ICDAR 2015\end{tabular}
& F1 ($\uparrow$)
& \begin{tabular}[t]{@{}l@{}}82.5 $\rightarrow$ \textbf{84.3} \\ 88.3 $\rightarrow$ \textbf{89.9}\end{tabular}
& Generating intermediate structured points via a two-stage prompt (SPOT). \\

\midrule
\multirow{17}{*}{\rotatebox{90}{\textbf{\textit{Layout}}}} 

& SCANBANK~\cite{kahu2021scanbank}
& \gdot{coloraug}
& ScanBank Test
& F1 Score ($\uparrow$)
& 45.9 $\rightarrow$ \textbf{53.5}
& Applying augmentations at source-code and image levels. \\

& READ~\cite{patil2020read}
& \gdot{colorgan}
& ICDAR2015
& mAP ($\uparrow$)
& 60.9 $\rightarrow$ \textbf{63.4}
& Generating layouts by sampling from a latent space. \\

& \cite{raman2022synthetic}
& \gdot{colorgan}
& PubLayNet
& F1 Score ($\uparrow$)
& 45.2 $\rightarrow$ \textbf{46.5}
& Generating layouts by sampling from a probabilistic model. \\

& \cite{pisaneschi2023automatic}
& \gdot{colorgan}
& \begin{tabular}[t]{@{}l@{}}ICPR 2021 \\ ICDAR 2019 \end{tabular}
& mAP ($\uparrow$)
& \begin{tabular}[t]{@{}l@{}}10.7 $\rightarrow$ \textbf{59.7} \\ 73.5 $\rightarrow$ \textbf{74.5} \end{tabular}
& Generating documents by populating transformer-based layouts. \\

& DIG~\cite{ying2024dig}
& \gdot{colorgan}
& \begin{tabular}[t]{@{}l@{}}DSSE-200 \\ PRIMA \end{tabular}
& mAP ($\uparrow$)
& \begin{tabular}[t]{@{}l@{}}47.05 $\rightarrow$ \textbf{56.07} \\ 53.80 $\rightarrow$ \textbf{62.26} \end{tabular}
& Generating realistic document images from a layout using a diffusion model. \\

& DocLayout-YOLO~\cite{zhao2024doclayout}
& \gdot{colorgan}
& \begin{tabular}[t]{@{}l@{}}DocLayNet \\ D4LA \end{tabular}
& mAP ($\uparrow$)
& \begin{tabular}[t]{@{}l@{}}76.7 $\rightarrow$ \textbf{79.3} \\ 68.6 $\rightarrow$ \textbf{69.8} \end{tabular}
& Synthesizing documents by iteratively packing elements onto a mesh. \\

& DLAgen~\cite{ying2024fine}
& \gdot{colorgan}
& Custom (ACL)
& mAP ($\uparrow$)
& 79.23 $\rightarrow$ \textbf{83.85}
& Synthesizing academic papers using context-free grammars. \\

& PubLayNet~\cite{zhong2019publaynet}
& \gdot{coloranno}
& SPD documents
& mAP ($\uparrow$)
& 62.0 $\rightarrow$ \textbf{66.3}
& Generating dataset via PDF-XML alignment. \\

& HJDataset~\cite{shen2020large}
& \gdot{coloranno}
& HJDataset
& mAP ($\uparrow$)
& 69.9 $\rightarrow$ \textbf{81.6}
& Annotating layouts via a semi-rule-based method. \\

& \cite{li2020cross}
& \gdot{coloranno}
& Legal to PubMed
& mAP ($\uparrow$)
& 68.6 $\rightarrow$ \textbf{70.7}
& Generating render-layer masks by parsing PDF metadata. \\

& \cite{ahuja2023new}
& \gdot{coloranno}
& Custom Test Set
& mAP ($\uparrow$)
& 83.2 $\rightarrow$ \textbf{88.6}
& Generating weak layout labels using YOLOv7. \\

& WordScape~\cite{weber2023wordscape}
& \gdot{coloranno}
& DocLayNet
& mAP ($\uparrow$)
& 29.9 $\rightarrow$ \textbf{50.8}
& Annotating layouts by rendering color-coded source files. \\

& MFCN~\cite{he2017multi}
& \gdot{colorssl}
& DSSE-200
& mean IoU ($\uparrow$)
& 73.3 $\rightarrow$ \textbf{75.9}
& Designing consistency and reconstruction pretext tasks. \\

& GenDoc~\cite{feng2023sequence}
& \gdot{colorssl}
& PubLayNet
& mAP ($\uparrow$)
& 90.15 $\rightarrow$ \textbf{92.31}
& Applying text infilling and coordinate prediction tasks. \\

\midrule
\multirow{3}{*}{\rotatebox{90}{\textbf{\textit{Table}}}} 

& CascadeTabNet~\cite{prasad2020cascadetabnet}
& \gdot{coloraug}
& ICDAR 2019
& F1 Score ($\uparrow$)
& 75.8 $\rightarrow$ \textbf{83.5}
& Applying dilation and smearing. \\

& TableBank~\cite{li2020tablebank}
& \gdot{coloranno}
& ICDAR 2013
& F1 Score ($\uparrow$)
& 87.86 $\rightarrow$ \textbf{96.25}
& Creating a dataset by parsing of Word/\LaTeX{} source. \\

& DocReL~\cite{li2022relational}
& \gdot{colorssl}
& SciTSR
& F1 ($\uparrow$)
& 98.65 $\rightarrow$ \textbf{99.78}
& Applying a relational consistency task. \\

\midrule
\multirow{2}{*}{\rotatebox{90}{\textbf{\textit{Order}}}} 

& LayoutReader~\cite{wang2021layoutreader}
& \gdot{coloranno}
& ReadingBank
& BLEU ($\uparrow$)
& 69.72 $\rightarrow$ \textbf{98.19}
& Extracting reading order from Word document XML. \\

& DocReL~\cite{li2022relational}
& \gdot{colorssl}
& ReadingBank
& BLEU ($\uparrow$)
& 97.38 $\rightarrow$ \textbf{98.41}
& Applying a relational consistency task. \\

\bottomrule
\end{tabularx}

\vspace{2pt}
\begin{minipage}{\linewidth}
\footnotesize
In the ``Type'' column, \gdot{coloraug} denotes Data Augmentation, \gdot{colorgan} is Data Generation from Scratch, \gdot{coloranno} is Automated Data Annotation, and \gdot{colorssl} is Self-Supervised Signal Construction. \textbf{Note: This color coding scheme is consistent across all tables in this section.}
\end{minipage}
\end{table}

\textbf{Performance on Text Recognition.}
As detailed in Table~\ref{tab:Document Parsing}, this cluster focuses on OCR and HTR. The primary bottleneck is the ``Style-Content Entanglement'': capturing the infinite variability of handwriting styles while maintaining diverse textual content is prohibitively expensive. To address this, generation methods aim to decouple these factors. \textbf{Data Generation from Scratch} has emerged as the dominant paradigm, diverging into two strategic paths. Deep Generative Models excel at synthesizing coherent global styles, delivering substantial gains (e.g., GC-DDPM~\cite{ding2023improving} reduces WER by 7.9\% on IAM), though they carry the risk of character hallucination. In contrast, Component-based Rendering guarantees content fidelity by stitching together real character crops, offering steady improvements (e.g., VATr~\cite{pippi2023handwritten} reduces WER by 2.8\% on Washington) despite potential artifacts in transitions. Notably, \textbf{Data Augmentation} remains a competitive baseline: FgAA~\cite{chen2025fine} achieves a 4.26\% WER reduction solely through stroke-level perturbations. This suggests that for fine-grained recognition, mastering local geometric distortions is as critical as modeling global style distributions, pointing toward future systems that synergize deep generative styles with precise structural control.

\textbf{Performance on Layout and Structure Analysis.}
Covering Layout Analysis, Table Detection, and Reading Order Detection (Table~\ref{tab:Document Parsing}), this cluster faces a dual challenge: the ``Long-tail Layout Distribution'' of real-world documents and the high cost of labeling complex logical structures. Consequently, \textbf{Data Generation from Scratch} serves as the primary engine for structural diversity. By synthesizing documents from randomized templates (e.g., DocLayout-YOLO~\cite{zhao2024doclayout}) or generative layout models (e.g., DIG~\cite{ying2024dig}), researchers can create infinite training samples with pixel-perfect ground truth, yielding robust gains (e.g., DIG improves mAP by 9.02\% on DSSE-200). Complementing this, \textbf{Automated Data Annotation} excels in acquiring precise logical labels for tasks like table recognition. Aligning PDF visual representations with source code (e.g., TableBank~\cite{li2020tablebank}, LayoutReader~\cite{wang2021layoutreader}) provides high-quality supervision where rule-based synthesis falls short, achieving dramatic improvements (e.g., LayoutReader boosts reading order BLEU by 28.47 points). Furthermore, \textbf{Self-Supervised Learning} is proving effective by forcing models to internalize the intrinsic ``grammar'' of document structures, delivering competitive improvements (GenDoc~\cite{feng2023sequence} improves mAP by 2.16\% on PubLayNet) through masked coordinate prediction tasks.

\textbf{Performance on Information Extraction.}
Encompassing Key Information Extraction, Named Entity Recognition, and Relation Extraction (Table~\ref{tab:Information Extraction}), this cluster demands a deep understanding of the ``Semantics-Layout Alignment''. The core bottleneck is the scarcity of documents annotated with fine-grained entity links. \textbf{Self-Supervised Signal Construction} emerges as the most influential paradigm here. By designing pretext tasks that mask both text and layout, models learn robust cross-modal representations, yielding consistent improvements across tasks (e.g., LayoutXLM~\cite{xu2021layoutxlm} boosts XFUND F1 by 8.11\%). \textbf{Automated Data Annotation} provides a scalable alternative. Techniques ranging from weak labeling with dictionaries to LLM-driven annotation generation (e.g., LayoutLLM~\cite{luo2024layoutllm}) have shown remarkable efficacy, with some methods (e.g., K2Q~\cite{zmigrod2024value}) nearly doubling performance on challenging benchmarks like DocILE. Finally, \textbf{Data Augmentation} remains vital for preventing overfitting on small datasets. Strategies that specifically perturb entity regions—such as token replacement (DADA~\cite{sun2024synonym}) or entity dropout~\cite{tang2023unifying}—force models to rely on context rather than memorization, delivering reliable gains (e.g., DADA improves FUNSD F1 by 0.28\%).

\begin{table}[t]
\caption{Performance Gains on \textbf{Information Extraction} Task.}
\label{tab:Information Extraction}
\scriptsize 
\renewcommand{\arraystretch}{1.3}
\begin{tabularx}{\textwidth}{
  c 
  >{\raggedright\arraybackslash}p{1.6cm}
  c 
  >{\raggedright\arraybackslash}p{1.5cm}
  >{\raggedright\arraybackslash}p{1.1cm}
  >{\raggedright\arraybackslash}p{1.3cm}
  >{\raggedright\arraybackslash}X 
}
\toprule
\textbf{Task} & \textbf{Key Work} & \textbf{Type} & \textbf{Benchmark} & \textbf{Metric} & \textbf{Gain} & \textbf{Key Contribution Highlight} \\
\midrule

\multirow{23}{*}{\rotatebox{90}{\textbf{\textit{KIE}}}} 

& \cite{shi2023multi}
& \gdot{coloraug}
& CORD
& F1 ($\uparrow$)
& 53.70 $\rightarrow$ \textbf{54.07}
& Simulating real-world noise by scaling text box content. \\

& \cite{dai2025enhancing}
& \gdot{coloraug}
& FormNLU
& mAP ($\uparrow$)
& 47.22 $\rightarrow$ \textbf{50.92}
& Simulating handwritten features using the Augraphy library. \\

& DADA~\cite{sun2024synonym}
& \gdot{coloraug}\gdot{coloranno}
& \begin{tabular}[t]{@{}l@{}}FUNSD \\ CORD\end{tabular}
& F1 ($\uparrow$)
& \begin{tabular}[t]{@{}l@{}}90.29 $\rightarrow$ \textbf{90.57} \\ 96.85 $\rightarrow$ \textbf{97.16}\end{tabular}
& Replacing text with GAN-generated images of their synonyms. \\

& EATEN~\cite{guo2019eaten}
& \gdot{colorgan}
& Train Ticket
& mEA ($\uparrow$)
& 86.2 $\rightarrow$ \textbf{95.8}
& Synthesizing data by populating templates with crawled text. \\

& NBID~\cite{wojcik2023nbid}
& \gdot{colorgan}
& Brazilian IDs
& F1 ($\uparrow$)
& 83.06 $\rightarrow$ \textbf{91.61}
& Rendering text onto templates created by GAN-based inpainting. \\

& Webvicob~\cite{kim2023web}
& \gdot{colorgan}
& \begin{tabular}[t]{@{}l@{}}FUNSD \\ XFUND\end{tabular}
& Acc. ($\uparrow$)
& \begin{tabular}[t]{@{}l@{}}60.59 $\rightarrow$ \textbf{73.97} \\ 78.17 $\rightarrow$ \textbf{84.55}\end{tabular}
& Extracting precise annotations from a rendered HTML DOM. \\

& DocILE\cite{vsimsa2023docile}
& \gdot{colorgan}\gdot{colorssl}
& DocILE Test
& F1 ($\uparrow$)
& 68.6 $\rightarrow$ \textbf{69.8}
& Combining rule-based synthesis with self-supervised MLM. \\

& LayoutLLM~\cite{luo2024layoutllm}
& \gdot{colorgan}\gdot{coloranno}\gdot{colorssl}
& FUNSD
& F1 ($\uparrow$)
& 70.96 $\rightarrow$ \textbf{79.98}
& Defining multi-level pre-training tasks and generating both LayoutCoT annotations and full HTML documents with an LLM. \\

& DAVID~\cite{ding2024david}
& \gdot{coloranno}
& FormNLU
& F1 ($\uparrow$)
& 85.76 $\rightarrow$ \textbf{88.61}
& Generating annotations using off-the-shelf tools and LLMs. \\

& K2Q~\cite{zmigrod2024value}
& \gdot{coloranno}
& \begin{tabular}[t]{@{}l@{}}DocILE \\ CORD\end{tabular}
& ANLS ($\uparrow$)
& \begin{tabular}[t]{@{}l@{}}53.0 $\rightarrow$ \textbf{90.0} \\ 43.6 $\rightarrow$ \textbf{94.0}\end{tabular}
& Generating training data by converting KIE datasets to a VQA format using templates. \\

& TGDOC~\cite{wang2023towards}
& \gdot{coloranno}\gdot{colorssl}
& FUNSD
& Acc. ($\uparrow$)
& 1.02 $\rightarrow$ \textbf{1.53}
& Generating instruction via an LLM and text spotting task. \\

& DocumentNet~\cite{yu2023documentnet}
& \gdot{coloranno}\gdot{colorssl}
& \begin{tabular}[t]{@{}l@{}}FUNSD \\ CORD\end{tabular}
& F1 ($\uparrow$)
& \begin{tabular}[t]{@{}l@{}}80.63 $\rightarrow$ \textbf{84.18} \\ 95.17 $\rightarrow$ \textbf{96.45}\end{tabular}
& Constructing self-supervised signals from weakly-annotated web pages. \\

& DocLayLLM~\cite{liao2025doclayllm}
& \gdot{coloranno}\gdot{colorssl}
& \begin{tabular}[t]{@{}l@{}}DeepForm \\ KLC\end{tabular}
& F1 ($\uparrow$)
& \begin{tabular}[t]{@{}l@{}}24.54 $\rightarrow$ \textbf{35.41} \\ 25.49 $\rightarrow$ \textbf{26.25}\end{tabular}
& Generating CoT instructions and defining self-supervised tasks. \\

& \cite{cao2023attention} 
& \gdot{colorssl}
& SROIE
& F1 ($\uparrow$)
& 59.3 $\rightarrow$ \textbf{85.8}
& Formulating pre-training tasks including Text-to-Segmentation. \\

& DocReL~\cite{li2022relational}
& \gdot{colorssl}
& FUNSD
& F1 ($\uparrow$)
& 42.86 $\rightarrow$ \textbf{46.08}
& Applying a self-supervised task based on relational consistency. \\

& ViT-VLP~\cite{mao2024visually}
& \gdot{colorssl}
& \begin{tabular}[t]{@{}l@{}}FUNSD \\ CORD\end{tabular}
& F1 ($\uparrow$)
& \begin{tabular}[t]{@{}l@{}}81.42 $\rightarrow$ \textbf{87.61} \\ 91.54 $\rightarrow$ \textbf{95.59}\end{tabular}
& Adding a generative layout modeling objective to pre-training. \\

\midrule
\multirow{3}{*}{\rotatebox{90}{\textbf{\textit{NER}}}} 

& \cite{Wojcik2025NewPI}
& \gdot{colorgan}\gdot{coloranno}
& \begin{tabular}[t]{@{}l@{}}FUNSD \\ EPHOIE\end{tabular}
& F1 ($\uparrow$)
& \begin{tabular}[t]{@{}l@{}}88.41 $\rightarrow$ \textbf{89.76} \\ 97.59 $\rightarrow$ \textbf{99.20}\end{tabular}
& Applying LLM-based rewriting and template-based document synthesis. \\

& LayoutXLM~\cite{xu2021layoutxlm}
& \gdot{colorssl}
& XFUND
& F1 ($\uparrow$)
& 74.71 $\rightarrow$ \textbf{82.82}
& Designing multimodal and multilingual pre-training tasks. \\

\midrule
\multirow{4}{*}{\rotatebox{90}{\textbf{\textit{RE}}}} 

& \cite{zhang2021entity}
& \gdot{coloraug}
& FUNSD
& F1 ($\uparrow$)
& 65.46 $\rightarrow$ \textbf{65.96}
& Augmenting data by randomly dropping words from entity phrases. \\

& \cite{Wojcik2025NewPI}
& \gdot{colorgan}\gdot{coloranno}
& FUNSD
& F1 ($\uparrow$)
& 62.76 $\rightarrow$ \textbf{70.52}
& Applying LLM-based rewriting and template-based document synthesis. \\

& \cite{zhang2021entity}
& \gdot{coloranno}
& FUNSD
& F1 ($\uparrow$)
& 60.35 $\rightarrow$ \textbf{63.81}
& Using auto-generated entity labels as additional features. \\

& LayoutXLM~\cite{xu2021layoutxlm}
& \gdot{colorssl}
& XFUND
& F1 ($\uparrow$)
& 60.02 $\rightarrow$ \textbf{72.06}
& Designing multimodal and multilingual pre-training tasks. \\

\bottomrule
\end{tabularx}
\end{table}

\begin{table}[t]
\caption{Performance Gains on \textbf{Document Understanding and Reasoning} Task.}
\label{tab:Document Understanding and Reasoning}
\scriptsize 
\renewcommand{\arraystretch}{1.3}
\begin{tabularx}{\textwidth}{
  c 
  >{\raggedright\arraybackslash}p{1.6cm}
  c 
  >{\raggedright\arraybackslash}p{1.1cm}
  >{\raggedright\arraybackslash}p{1.3cm}
  >{\raggedright\arraybackslash}p{1.3cm}
  >{\raggedright\arraybackslash}X 
}
\toprule
\textbf{Task} & \textbf{Key Work} & \textbf{Type} & \textbf{Benchmark} & \textbf{Metric} & \textbf{Gain} & \textbf{Key Contribution Highlight} \\
\midrule
\multirow{5}{*}{\rotatebox{90}{\textbf{\textit{Classification}}}} 

& DocumentNet~\cite{yu2023documentnet}
& \gdot{coloranno}\gdot{colorssl}
& RVL-CDIP
& Acc. ($\uparrow$)
& 93.47 $\rightarrow$ \textbf{95.34}
& Constructing self-supervised signals from weakly-annotated web pages. \\

& DiT~\cite{li2022dit}
& \gdot{colorssl}
& RVL-CDIP
& Acc. ($\uparrow$)
& 91.11 $\rightarrow$ \textbf{92.69}
& Constructing visual token prediction tasks using a doc-specific dVAE. \\

& Donut~\cite{kim2022ocr}
& \gdot{colorssl}
& RVL-CDIP
& Acc. ($\uparrow$)
& 94.42 $\rightarrow$ \textbf{95.30}
& Establishing an OCR-free baseline surpassing OCR-dependent models. \\

& LayoutLMv3~\cite{huang2022layoutlmv3}
& \gdot{colorssl}
& RVL-CDIP
& Acc. ($\uparrow$)
& 95.05 $\rightarrow$ \textbf{95.44}
& Unified pre-training with MIM and MLM. \\

& UDOP~\cite{tang2023unifying}
& \gdot{colorssl}
& RVL-CDIP
& Acc. ($\uparrow$)
& 95.3 $\rightarrow$ \textbf{96.2}
& Designing joint text-layout reconstruction self-supervised task. \\

\midrule
\multirow{9}{*}{\rotatebox{90}{\textbf{\textit{Chart Understanding}}}} 

& ChartBench~\cite{xu2023chartbench}
& \gdot{colorgan}
& ChartBench
& Acc. ($\uparrow$)
& 26.98 $\rightarrow$ \textbf{39.99}
& Synthesizing chart QA pairs from rendered, LLM-generated data. \\

& StructChart~\cite{xia2023structchart}
& \gdot{colorgan}
& ChartQA
& mPrecision ($\uparrow$)
& 67.70 $\rightarrow$ \textbf{71.16}
& Generating chart images by executing LLM-written code. \\

& ChartOCR~\cite{luo2021chartocr}
& \gdot{coloranno} 
& FQA (Bar)
& Mean Error ($\downarrow$)
& 50.0 $\rightarrow$ \textbf{18.5}
& Generating chart-and-table pairs by rendering spreadsheets. \\

& MMC~\cite{liu2024mmc}
& \gdot{coloranno}
& ChartQA
& Acc. ($\uparrow$)
& 51.6 $\rightarrow$ \textbf{57.4}
& Generating question-answer pairs for charts using GPT-4. \\

& mPLUG-DocOwl 1.5~\cite{hu2024mplug}
& \gdot{coloranno}\gdot{colorssl}
& \begin{tabular}[t]{@{}l@{}}ChartQA \\ TabFact\end{tabular}
& Acc. ($\uparrow$)
& \begin{tabular}[t]{@{}l@{}}65.0 $\rightarrow$ \textbf{67.5} \\ 72.9 $\rightarrow$ \textbf{76.5}\end{tabular}
& Combining self-supervised structure parsing with LLM-based explanation generation. \\

& Mono-InternVL~\cite{luo2025mono}
& \gdot{colorssl}
& \begin{tabular}[t]{@{}l@{}}ChartQA \\ AI2D\end{tabular}
& Acc. ($\uparrow$)
& \begin{tabular}[t]{@{}l@{}}11.1 $\rightarrow$ \textbf{13.5} \\ 26.6 $\rightarrow$ \textbf{42.7}\end{tabular}
& Generating training signals via a multi-stage auto-regressive objective. \\

& UniChart~\cite{masry2023unichart}
& \gdot{colorssl}
& ChartQA
& Acc. ($\uparrow$)
& 59.76 $\rightarrow$ \textbf{64.72}
& Designing chart-specific pre-training for multi-level understanding. \\

\midrule
\multirow{25}{*}{\rotatebox{90}{\textbf{\textit{DocVQA}}}} 

& TILT~\cite{powalski2021going}
& \gdot{coloraug}
& DocVQA
& ANLS ($\uparrow$)
& 82.2 $\rightarrow$ \textbf{82.9}
& Applying a simple case-alternation data augmentation. \\

& LATIN-Tuning~\cite{wang2023layout}
& \gdot{colorgan}
& \begin{tabular}[t]{@{}l@{}}DocVQA \\ InfoVQA\end{tabular}
& ANLS ($\uparrow$)
& \begin{tabular}[t]{@{}l@{}}35.67 $\rightarrow$ \textbf{66.97} \\ 14.19 $\rightarrow$ \textbf{30.28}\end{tabular}
& Generating spatially-aware instruction from tables using an LLM. \\

& Webvicob~\cite{kim2023web}
& \gdot{colorgan}
& DocVQA
& ANLS ($\uparrow$)
& 44.77 $\rightarrow$ \textbf{56.07}
& Extracting precise annotations from a rendered HTML DOM. \\

& TextSquare~\cite{tang2024textsquare}
& \gdot{colorgan}
& DocVQA 
& ANLS ($\uparrow$)
& 74.8 $\rightarrow$ \textbf{84.3}
& Generating QA-reasoning data via a questioning and evaluation loop. \\

& DocKD~\cite{kim2024dockd}
& \gdot{colorgan}\gdot{coloranno}
& DocVQA
& ANLS ($\uparrow$)
& 80.6 $\rightarrow$ \textbf{83.4}
& Generating annotations via LLM-prompting and text-rendering. \\

& LayoutLLM~\cite{luo2024layoutllm}
& \gdot{colorgan}\gdot{coloranno}\gdot{colorssl}
& DocVQA
& ANLS ($\uparrow$)
& 70.82 $\rightarrow$ \textbf{74.27}
& Defining multi-level pre-training tasks and generating both LayoutCoT annotations and full HTML documents with an LLM. \\

& \cite{chen2024expanding}
& \gdot{coloranno}
& DocVQA
& ANLS ($\uparrow$)
& 94.1 $\rightarrow$ \textbf{95.1}
& Filtering data using LLM-generated quality scores. \\

& \cite{laurenccon2024building}
& \gdot{coloranno}
& DocVQA
& ANLS ($\uparrow$)
& 60.1 $\rightarrow$ \textbf{71.4}
& Generating QA pairs from document text using an LLM. \\

& LLaVAR~\cite{zhang2023llavar}
& \gdot{coloranno}
& DocVQA
& Acc. ($\uparrow$)
& 6.9 $\rightarrow$ \textbf{11.6}
& Generating grounded, multi-turn dialogues using GPT-4. \\

& Meteor~\cite{lee2024meteor}
& \gdot{coloranno}
& \begin{tabular}[t]{@{}l@{}}MMBench \\ MM-Vet\end{tabular}
& Acc. ($\uparrow$)
& \begin{tabular}[t]{@{}l@{}}73.2 $\rightarrow$ \textbf{77.0} \\ 44.8 $\rightarrow$ \textbf{48.2}\end{tabular}
& Generating detailed rationales for existing QA pairs. \\

& \cite{li2024multimodal}
& \gdot{coloranno}
& MathVista
& Acc. ($\uparrow$)
& 40.0 $\rightarrow$ \textbf{50.2}
& Generating QA pairs from scientific figures using GPT-4V. \\

& Oasis~\cite{zhang2025oasis}
& \gdot{coloranno}
& \begin{tabular}[t]{@{}l@{}}DocVQA \\ InfoVQA\end{tabular}
& \begin{tabular}[t]{@{}l@{}}ANLS ($\uparrow$) \\ Acc. ($\uparrow$)\end{tabular}
& \begin{tabular}[t]{@{}l@{}}71.7 $\rightarrow$ \textbf{76.0} \\ 33.3 $\rightarrow$ \textbf{39.6}\end{tabular}
& Generating instruction from images alone using a MLLM. \\

& UniDoc~\cite{feng2023unidoc}
& \gdot{coloranno}\gdot{colorssl}
& DocVQA
& Acc. ($\uparrow$)
& 35.78 $\rightarrow$ \textbf{40.72}
& Generating instruction by creating dialogues and OCR tasks. \\

& DocLayLLM~\cite{liao2025doclayllm}
& \gdot{coloranno}\gdot{colorssl}
& DocVQA
& ANLS ($\uparrow$)
& 76.00 $\rightarrow$ \textbf{77.43}
& Generating CoT instructions and defining self-supervised tasks. \\

& MarkupLM~\cite{li2022markuplm}
& \gdot{colorssl}
& WebSRC
& EM ($\uparrow$)
& 54.29 $\rightarrow$ \textbf{59.56}
& Designing hierarchical pre-training tasks based on HTML. \\

& DocPedia~\cite{feng2024docpedia}
& \gdot{colorssl}
& DocVQA
& Acc. ($\uparrow$)
& 23.0 $\rightarrow$ \textbf{39.5}
& Formulating a suite of self-supervised text perception tasks. \\

& DoCo~\cite{li2024enhancing}
& \gdot{colorssl}
& DocVQA
& Acc. ($\uparrow$)
& 62.2 $\rightarrow$ \textbf{64.8}
& Defining an object-level contrastive learning task. \\

& MATCHA~\cite{liu2023matcha}
& \gdot{colorssl}
& DocVQA
& Acc. ($\uparrow$)
& 72.1 $\rightarrow$ \textbf{74.2}
& Designing chart de-rendering and math reasoning pre-training tasks. \\

& Pix2Struct~\cite{lee2023pix2struct}
& \gdot{colorssl}
& DocVQA
& ANLS ($\uparrow$)
& 12.2 $\rightarrow$ \textbf{67.8}
& Designing a screenshot parsing (image-to-HTML) pre-training task. \\

& GenDoc~\cite{feng2023sequence}
& \gdot{colorssl}
& DocVQA
& ANLS ($\uparrow$)
& 64.87 $\rightarrow$ \textbf{78.15}
& Applying text infilling and coordinate prediction pre-training tasks. \\

& UDOP~\cite{tang2023unifying}
& \gdot{colorssl}
& DocVQA
& ANLS ($\uparrow$)
& 79.7 $\rightarrow$ \textbf{84.4}
& Designing joint text-layout reconstruction pre-training tasks. \\

\bottomrule
\end{tabularx}
\end{table}

\textbf{Performance on Document Classification.}
As summarized in Table~\ref{tab:Document Understanding and Reasoning}, the primary challenge is learning discriminative ``Global Document Representations'' that are robust to layout variations. Consequently, \textbf{Self-Supervised Signal Construction} dominates this landscape. Pre-training objectives that enforce global consistency—such as joint text-layout reconstruction (UDOP~\cite{tang2023unifying}) or cross-modal alignment (DocumentNet~\cite{yu2023documentnet})—empower models to learn powerful document embeddings from abundant unlabeled data, resulting in steady gains (e.g., UDOP improves RVL-CDIP accuracy by 0.9\% over strong baselines). While less prevalent, \textbf{Data Generation from Scratch} also contributes by synthesizing diverse document templates to enrich the training distribution.

\textbf{Performance on Chart Understanding.}
Detailed in Table~\ref{tab:Document Understanding and Reasoning}, this cluster tackles the challenge of ``Data-Visual Reverse Engineering'': translating abstract visual cues into numerical logic. The bottleneck is the scarcity of high-quality chart-QA pairs. \textbf{Automated Data Annotation} leads the solution space. Whether by prompting LLMs for direct reasoning or de-rendering charts into intermediate data tables, this paradigm delivers substantial gains. For instance, ChartOCR~\cite{luo2021chartocr} reduces mean relative error by 31.5\% on FQA. Complementing this, \textbf{Data Generation from Scratch} builds foundational robustness. By rendering millions of synthetic chart-table pairs via plotting libraries, methods like UniChart~\cite{masry2023unichart} enable effective pre-training, yielding a 4.96\% accuracy boost on ChartQA.

\textbf{Performance on Document Visual Question Answering.}
As the pinnacle of DI, detailed in Table~\ref{tab:Document Understanding and Reasoning}, this task demands ``Cross-modal Complex Reasoning''. The primary limitation of existing datasets is the lack of explicit reasoning chains; models often learn to extract answers without understanding the underlying logic. \textbf{Automated Data Annotation} is the dominant force in bridging this cognitive gap. By prompting LLMs to synthesize not just answers but also detailed rationales and instruction-following traces, researchers have unlocked significant performance leaps. For example, LLaVAR~\cite{zhang2023llavar} improves DocVQA accuracy by 4.7\% solely through LLM-generated instruction tuning. \textbf{Self-Supervised Signal Construction} provides the necessary pre-training foundation. Methods like UDOP~\cite{tang2023unifying}, which unify vision, text, and layout generation, demonstrate that mastering holistic document context yields significant dividends, boosting ANLS by 4.7 points.

\subsection{Challenges in Data Evaluation}
\label{sec:eval_challenges}
While the proposed metrics provide a foundation, accurately assessing generated data remains fraught with difficulties.

First is the \textbf{Dataset Shortcuts}. Models may achieve high scores by exploiting ``shortcuts'' rather than mastering a generalizable capability. This manifests in two main ways: template leakage, where models ``memorize'' templates shared between train and test sets~\cite{laatiri2023information}, and modality redundancy, where samples can be answered without referencing the image~\cite{chen2024we}. Detecting such shortcuts is essential to ensure that reported performance gains (Section ~\ref{sec:extrinsic_eval}) are genuine.

Second is the \textbf{Alignment with Human Perception}. A significant gap exists between current objective metrics and human judgments on aesthetics and functionality. As noted by Uni-Layout~\cite{lu2025uni}, a layout that scores highly on geometric metrics may still appear ugly or impractical to human evaluators. Developing automated metrics that can capture these subtle aesthetic nuances and better align with comprehensive human perception remains an open problem.

Finally, there is the need for \textbf{In-depth Evaluation of Complex Generated Content}. As LLMs can generate complex content like instructions, question-answers, and reasoning chains (Section~\ref{sec:llm_annotation}), evaluation must shift from simple ``right or wrong'' judgments to a deeper quality assessment. Future evaluations need to assess abstract dimensions like solvability, clarity, hallucinations, and logical complexity, as pointed out by Oasis~\cite{zhang2025oasis} and ProcTag~\cite{shen2025proctag}.

\section{Insights and Evolving Trends}
\label{sec:discussion}
The preceding sections have systematically dissected the methodologies (Sections~\ref{sec:data_augmentation}–\ref{sec:self_supervised}) and evaluation protocols (Section~\ref{sec:evaluation}) of data generation in DI. Stepping back from these technical details, we now examine the broader data production ecosystem from a macroscopic perspective. In this section, we distill the core evolutionary trends shaping the future of data production. We organize these insights hierarchically: starting from the Technological Cornerstones that power the revolution (Section~\ref{sec:cornerstones}), identifying the profound Paradigm Shifts in objectives and processes (Section~\ref{sec:shifts}), exploring the Technical Frontiers required to realize these shifts (Section~\ref{sec:frontiers}), and finally envisioning the ultimate Ecosystem Evolution towards a self-sustaining co-evolutionary loop (Section~\ref{sec:evolution}).

\subsection{Technological Cornerstones}
\label{sec:cornerstones}
We first observe two cornerstones that form the engine and foundation of the current data production revolution. The first is \textbf{Generative AI has become a Universal, Cross-Paradigm Engine}. Traditionally, different data production tasks relied on specialized tech stacks. However, generative AI, particularly LLMs, is breaking down these barriers, evolving into a universal core technology that can be flexibly invoked to empower all stages of data production. Second, \textbf{Self-Supervised Pre-training has become the Universal Infrastructure} that underpins this engine. Crucially, self-supervised learning (Section~\ref{sec:self_supervised}) is itself the most foundational form of data generation, creating pretext tasks from unlabeled data to produce powerful pre-trained models. This model then becomes the digital infrastructure that enables all other data production paradigms to operate efficiently: it can be directly used as a pre-labeler for automated annotation (Section~\ref{sec:auto_annotation}), or serve as a powerful backbone for generation from scratch (Section~\ref{sec:generation_from_scratch}). Therefore, self-supervised pre-training is not an isolated step but the very starting point for the entire data production chain.

Built upon these foundations, we advocate for a \textbf{Strategic Hybridization} of paradigms. Rather than selecting a single method in isolation, a robust pipeline should orchestrate them synergistically: employing \textit{Data Generation from Scratch} to cover long-tail distributions, leveraging \textit{Automated Annotation} to ensure precision for head distributions, and finally applying \textit{Data Augmentation} to enforce local robustness. This strategic combination effectively bridges the gap between theoretical capability and practical deployment, maximizing the utility of the universal engine and infrastructure. Ultimately, this hybridization unifies the resource-centric decision logic (Section~\ref{sec:taxonomy}) with task-specific performance demands (Section~\ref{sec:evaluation}), offering a blueprint for navigating the diverse landscape of data generation.

\subsection{Paradigm Shifts}
\label{sec:shifts}
Based on these cornerstones, the core objectives and processes of data production are undergoing two profound shifts. The first is an \textbf{objective shift from ``Representation Learning'' to ``Knowledge Distillation.''} The ultimate goal of data generation is no longer just to learn representations, but to distill the implicit knowledge from powerful teacher models into explicit, high-quality training data. For instance, prompting an LLM for chain-of-thought annotations or self-question-and-answer routines are key mechanisms for extracting these ``knowledge crystals.'' This reframes data generation as a form of knowledge engineering aimed at training smaller, specialized models. The second is a \textbf{process shift from ``Static Data Pools'' to ``Dynamic Curricula.''} The focus is shifting from data quantity to intelligent orchestration of its sequence and difficulty. By integrating data generation with curriculum learning, we can tailor optimal learning paths for models. To realize this, the community should focus on developing metadata-driven curriculum generators that explicitly tag generated samples with complexity attributes (e.g., layout density) to mathematically optimize the training schedule. This empowers systems to diagnose knowledge gaps and generate personalized textbooks on demand.

\subsection{Technical Frontiers}
\label{sec:frontiers}
To support these paradigm shifts, we identify three promising technical frontiers that warrant further exploration.
First, data generation is evolving from static workflows to \textbf{agentic generation}. Future systems will employ autonomous agents capable of dynamically planning strategies\textemdash searching for templates, simulating interactions, and verifying outputs. While this flexibility addresses the rigidity of current pipelines, it necessitates robust error-recovery mechanisms to prevent cascading failures in autonomous loops.
Second, to mitigate hallucinations in domain-specific tasks, researchers are prioritizing \textbf{knowledge injection}. Integrating external databases via Retrieval-Augmented Generation (RAG) allows models to condition synthesis on retrieved facts, ensuring documents are both visually realistic and factually grounded. However, this introduces new complexities in aligning multimodal retrieval with generative objectives.
Finally, the role of humans is shifting from annotators to supervisors through \textbf{collaborative creation}. Emerging workflows leverage Reinforcement Learning from Human Feedback (RLHF), where experts provide high-level critiques rather than instance-level labels. This co-creation paradigm maximizes data utility with minimal effort, though it demands careful reward modeling to avoid reward hacking.

\subsection{Ecosystem Evolution}
\label{sec:evolution}
Finally, guided by these new objectives and processes, the ecosystem's driving force and ultimate form are evolving. The first is a \textbf{driving force shift from generation-centric to evaluation-centric.} The key question is shifting from ``What can we generate?'' to ``What is actually useful?''. A closed loop between generation and evaluation, driven by advanced evaluation methods, is forming. As demonstrated by systems that use reinforcement learning to optimize generators or incorporate a self-evaluation step, evaluation is no longer the end point but the starting point for optimization. This suggests that the most powerful data generation systems will be self-evolving, comprising a generator and a powerful, objective-aligned evaluator. Propelled by this force, the ecosystem is undergoing a \textbf{form shift from ``Integrated Workflows'' to a ``Co-evolutionary Ecosystem.''} The ultimate form is evolving from a linear pipeline into a recursive, self-improving cycle: models learn from data, generate better data, and are then improved by that data. Concepts like self-iteration foreshadow future AI with autonomous growth, where data generation shifts from a one-time feeding to the critical fuel sustaining this bootstrapping cycle.

However, this recursive ecosystem faces a fundamental existential threat: \textbf{Data Entropy.} Without intervention, the minor hallucinations inherent in generative models, fidelity gaps in synthetic renderings, and biases amplified by automation will accumulate over iterations, causing the system to degenerate—a phenomenon known as Model Collapse. Therefore, the longevity of this ecosystem depends not just on generation speed, but on implementing \textbf{Negentropy Filters}---rigorous, automated auditing mechanisms (as discussed in Section~\ref{sec:evaluation}) that continuously purify the training stream, ensuring that the self-improving cycle remains constructive rather than degenerative.

\section{CONCLUSION}
\label{sec:conclusion}
This survey has provided the first systematic exploration of Data Generation methods in Document Intelligence, proposing a novel taxonomy centered on \textbf{resource constraints and learning objectives}. Guided by this framework, we have dissected the methodological landscape across four paradigms: identifying how \textit{Data Augmentation} enhances robustness via perturbations; how \textit{Data Generation from Scratch} bridges the cold-start gap through template rendering and generative AI; how \textit{Automated Data Annotation} accelerates labeling by shifting from heuristic rules to LLM-driven reasoning; and how \textit{Self-Supervised Signal Construction} mines intrinsic supervision to empower foundation models. 
Furthermore, we established a multi-level evaluation framework integrating intrinsic quality and extrinsic utility. Synthesizing these findings, we distilled the field's trajectory toward a new ecosystem driven by generative AI, centered on evaluation, and characterized by the co-evolution of data and models. In conclusion, Data Generation methods are no longer supplementary tools but are forming a new, self-improving paradigm of data production. The ability to apply these methods will be the decisive force in creating the next generation of document intelligence systems.

\bibliographystyle{ACM-Reference-Format}
\bibliography{survey}

\end{document}